\documentclass[ai,article,submit,pdftex,moreauthors]{Definitions/mdpi} 
\renewcommand{\linenumbers}{}

\firstpage{1} 
\pubvolume{1}
\issuenum{1}
\articlenumber{0}
\pubyear{2024}
\copyrightyear{2024}
\datereceived{ } 
\daterevised{ } 
\dateaccepted{ } 
\datepublished{ } 
\hreflink{https://doi.org/} 

\usepackage[draft, xcolor=dvipsnames]{changes}

\definechangesauthor[name=WK, color=blue]{1}
\definechangesauthor[name=RE, color=olive]{2}
\definechangesauthor[name=MO, color=red]{3}
\definechangesauthor[name=RD, color=orange]{4}
\definechangesauthor[name=GR, color=lime]{5}
\definechangesauthor[name=REV, color=cyan]{6}
\setauthormarkuptext{name} 

\newcommand{\noshow}[1]{}
 
\usepackage{siunitx}
\usepackage{tikz}
\usetikzlibrary{positioning, fit, calc, shapes.multipart, arrows.meta}
\definecolor{keywordcolor}{rgb}{0.13,0.29,0.53}
\definecolor{stringcolor}{rgb}{0.31,0.60,0.02}

\Title{ChatGPT Code Detection: Techniques for Uncovering the Source of Code}

\TitleCitation{ChatGPT Code Detection: Techniques for Uncovering the Source of Code}

\Author{Marc Oedingen $^{1}$\orcidA{}, Raphael C. Engelhardt $^{1}$\orcidB{}, Robin Denz $^{2}$\orcidC{}, Maximilian Hammer $^{1}$\orcidD{} and Wolfgang Konen $^{1}$\orcidE{}*}

\AuthorNames{Marc Oedingen; Raphael C. Engelhardt; Robin Denz; Maximilian Hammer and Wolfgang Konen}

\AuthorCitation{Oedingen, M.; Engelhardt, R. C.; Denz, R.; Hammer M.; Konen, W.}

\address{%
$^{1}$ \quad Cologne Institute of Computer Science, Faculty of Computer Science and Engineering Science, TH Köln, Gummersbach, Germany; \{Marc.Oedingen,Raphael.Engelhardt,Wolfgang.Konen\}@th-koeln.de and Maximilian.Hammer@smail.th-koeln.de\\
$^{2}$ \quad Department of Medical Informatics, Biometry and Epidemiology, Faculty of Medicine,  Ruhr-University Bochum, Bochum, Germany; denz@amib.rub.de}

\corres{Correspondence: Wolfgang.konen@th-koeln.de
}

\abstract{
In recent times, large language models (LLMs) have made significant strides in generating computer code, blurring the lines between code created by humans and code produced by artificial intelligence (AI). As these technologies evolve rapidly, it is crucial to explore how they influence code generation, especially given the risk of misuse in areas like higher education. This paper explores this issue by using advanced classification techniques to differentiate between code written by humans and that generated by ChatGPT, a type of LLM. We employ a new approach that combines powerful embedding features (black-box) with supervised learning algorithms -- including Deep Neural Networks, Random Forests, and Extreme Gradient Boosting -- to achieve this differentiation with an impressive accuracy of $98\%$. 
For the successful combinations, we also examine their model calibration, showing that some of the models are extremely well calibrated. Additionally, we present white-box features and an interpretable Bayes classifier to elucidate critical differences between the code sources, enhancing the explainability and transparency of our approach. Both approaches work well but provide at most $85-88\%$ accuracy. We also show that untrained humans solve the same task not better than random guessing. This study is crucial in understanding and mitigating the potential risks associated with using AI in code generation, particularly in the context of higher education, software development, and competitive programming. 
}


\begin{document}


\section{Introduction}
\label{sec:introduction}

The recent performance of ChatGPT has astonished scholars and the general public, not only regarding the seemingly human way of using natural language but also its proficiency in programming languages. While the training data (e.g., for the Python programming language) ultimately stems from human programmers, ChatGPT has likely developed its own coding style and idiosyncrasies, as human programmers do. In this paper, we train and evaluate various machine learning (\textbf{ML}) models on distinguishing human-written from ChatGPT-written Python code. These models achieve very high performance even for code samples with few lines, a seemingly impossible task for humans.
The following subsections present our motivation, scientific description of the task, and the elaboration of research questions.

\subsection{Motivation}
\label{sec:motivation}
Presently, the world is looking ambivalently at the development and opportunities of powerful large language models (\textbf{LLMs}). On the one hand, such models can execute complex tasks and augment human productivity due to their enhanced performance in various areas \cite{alawida2023comprehensive, ziegler2022productivity}. On the other hand, these models can be misused for malicious purposes, such as generating deceptive articles or cheating in educational institutions and other competitive environments \cite{charan2023text, letterAI, PolicyAI, weidinger2021ethical, zhang2023ethical}.

The inherently opaque nature of these black-box LLM models, combined with the difficulty of distinguishing between human- and AI-generated content, poses a problem that can make it challenging to trust these models \cite{ethics}. Earlier research has extensively focused on detecting natural language (\textbf{NL}) text content generated by LLMs \cite{mitchell2023detectgpt, gehrmann2019gltr, alamleh2023distinguishing}. A recent survey~\cite{ghosal2023possibilities} discusses the strengths and weaknesses of those approaches.
%
However, detecting AI-generated code is an equally important and relatively unexplored area of research. As LLMs are utilized more often in the field of software development, the ability to distinguish between human- and AI-generated code becomes increasingly important. 
Thereby, it is not exclusively about the distinction but also the implications, such as the code's trustworthiness, efficiency, and security. Moreover, the rapid development and improvement of AI models may lead to an arms race where detection techniques must continuously evolve to stay ahead of the curve.

Earlier research showed that using LLMs to generate code 
can lead to security vulnerabilities, and $40\%$ of the code fails to solve the given task \cite{pearce2021asleep}. Contrastingly, using LLMs can also lead to a significant increase in productivity and efficiency \cite{ziegler2022productivity}. 
This dual-edged nature of LLMs necessitates a balanced approach. Harnessing the potential benefits of such models while mitigating risks is the key. Ensuring the authenticity of code is especially crucial in academic environments, where the integrity of research and educational outcomes is paramount. Fraudulent or AI-generated submissions can undermine the foundation of academic pursuits, leading to a loss of trust in research findings and educational qualifications. Moreover, in the context of examinations, robust fraud detection is essential to prevent cheating, ensuring the assessments accurately reflect the student's capabilities and do not check the non-deterministic output of a prompt due to the stochastic decision-making of LLMs based on transformer models (\textbf{TM}) during inference. Under the assumption that AI-generated code has a higher chance of security vulnerabilities and beyond the educational context, it can also be critical in software industries to have the ability to distinguish between human- and AI-generated code when testing an unknown piece of software. As the boundaries of what AI can achieve expand, our approach to understanding, managing, and integrating these capabilities into our societal fabric, including academic settings, will determine our success in the AI-augmented era.

\subsection{Problem Introduction}
\label{sec:problem_intro}

This paper delves deep into the challenge of distinguishing between human-generated and AI-generated code, offering a comprehensive overview of state-of-the-art methods and proposing novel strategies to tackle this problem.

\begin{figure}[h]
    \centering
    \input{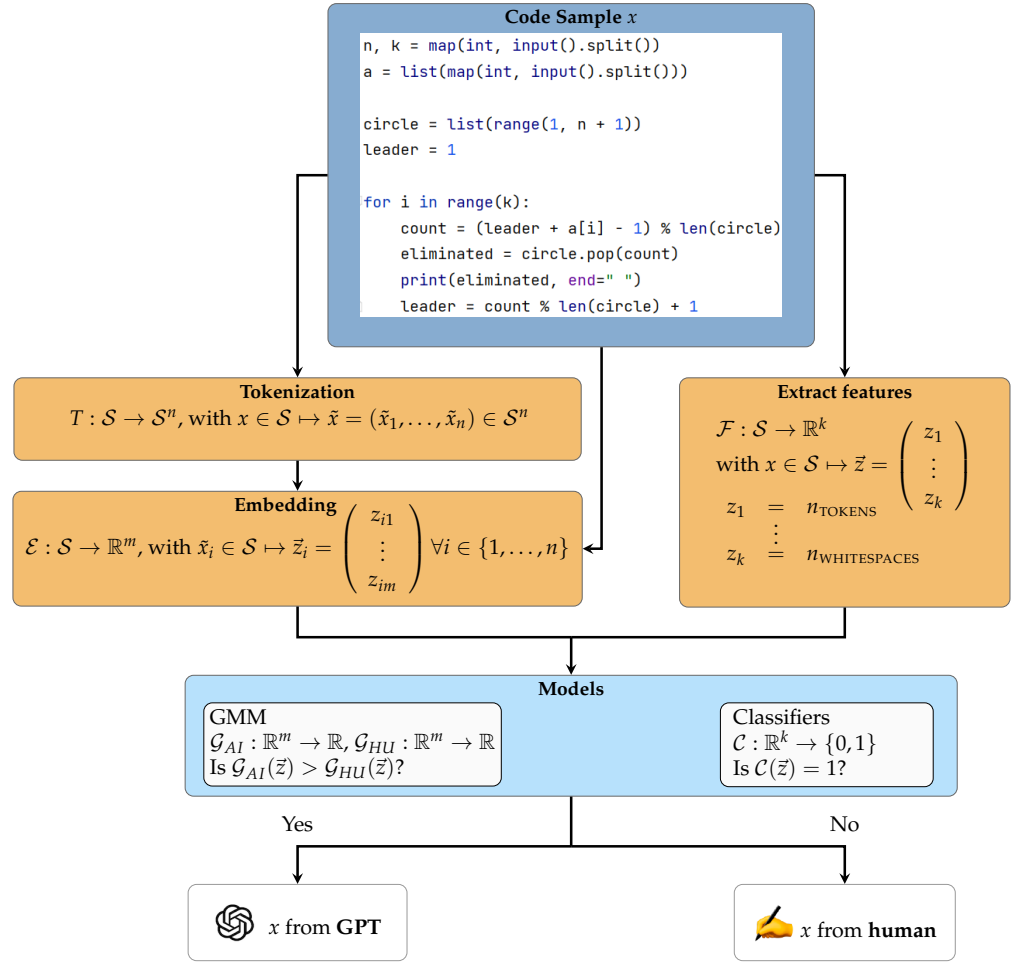}
    \caption{Flowchart of our Code Detection Methodology 
    }
    \label{fig:CD_flow_chart}
\end{figure}

Central to our methodology is a reduction of complexity: the intricate task of differentiating between human- and AI-generated code
is represented as a fundamental binary classification problem. Specifically, given a code snippet $x \in \mathcal{C}$ as input, we aim for a function $f : \mathcal{C} \rightarrow \{0,1\} = \mathcal{Y}$, which indicates whether the code's origin is human $\{0\}$ or GPT $\{1\}$. This allows us to use well-known and established ML models. We represent the code snippets as human-designed (white-box) and embeddings (black-box) features in order to apply ML models. The usage of embeddings requires prior tokenization, which is carried out either implicitly by the model or explicitly by us. 
Hence, we use embeddings to obtain constant dimensionality across all code snippets or single tokens. 
Figure~\ref{fig:CD_flow_chart} gives an overview of our approach in the form of a flowchart, where the details will be explained in the following sections. 

For a model to truly generalize in the huge field of software development across many tasks, it requires training on vast amounts of data. However, volume alone is insufficient for a model. The data's quality is paramount, ensuring meaningful and discriminating features are present within the code snippets. An ideal dataset would consist of code snippets that satisfy a variety of test cases to guarantee their syntactic and semantic correctness for a given task. Moreover, to overcome the pitfalls of biased data, we emphasize snippets prior to the proliferation of GPT in code generation. However, a significant blocker emerges from a stark paucity of publicly available GPT-generated solutions that match our criteria mentioned before. This lack underscores the necessity to find and generate solutions that can serve as fitting data sources for our classifiers.

\subsection{Research Questions and Contributions}
We formulate the following research questions based on the problem introduction and the need to obtain detection techniques for AI-generated code. In addition, we provide our hypotheses regarding the questions we want to answer with this work.

\begin{quote} \label{RQ:1}
    \hspace*{-0.75cm}
    \emph{\textbf{RQ1}: Can we distinguish (on unseen tasks) between human- and AI-generated code?}
\end{quote}

\begin{quote} \label{RQ:2}
    \hspace*{-0.75cm}
    \emph{\textbf{RQ2}: To what extent can we explain the difference between human- and AI-generated code?}
\end{quote}

\begin{itemize}
    \item \emph{$\mathbf{H_1}$}: There are detectable differences in style between AI-generated and human-generated code. 
    \item \emph{$\mathbf{H_2}$}: The differences are only to a minor extent attributable to code formatting: If both code snippets are formatted in the same way, there are still many detectable differences.
\end{itemize}

Upon rigorous scrutiny of the posed research questions, an apparent paradox emerges. LLMs have been trained using human-generated code. Consequently, the question arises: Does AI-generated code diverge from its human counterpart? We postulate that LLMs follow learning trajectories similar to individual humans. Throughout the learning process, humans and machines are exposed to many code snippets, each encompassing distinct stylistic elements. They subsequently develop and refine their unique coding style, i.e., by using consistent variable naming conventions, commenting patterns, code formatting, or selecting specific algorithms for given scenarios \cite{yasir2023exploring}. Given the vast amount of code the machine has seen during training, it is anticipated to adopt a more generalized coding style. Thus, identifiable discrepancies between machine-generated code snippets and individual human-authored code are to be expected.

The main contributions of this paper are: 
\begin{enumerate}
    \item Several classification models are evaluated on a large corpus of code data.
    While the human-generated code comes from many different subjects, the AI-generated code is (currently) only produced by ChatGPT-3.5.
    \item The best model-feature combinations are models operating on high dimensional vector embeddings (black-box) of the code data. 
    \item Formatting all snippets with the same code formatter decreases the accuracy only slightly. Thus, the format of the code is \textit{not} the key feature of distinction.
    \item The best models achieve classification accuracies of $98\%$. 
    An explainable classifier with almost $90\%$ accuracy is obtained with the help of Bayes classification. 
\end{enumerate}

\subsection{Structure}
The remainder of this article is structured as follows: Section~\ref{sec:related} provides a comprehensive review of the existing literature, evaluating and discussing its current state. Section~\ref{sec:methodology} offers an overview of the techniques and frameworks employed throughout this study, laying the foundation for understanding the subsequent sections. Section~\ref{sec:exp_setup} details our experimental setup, outlining procedures from data collection and preprocessing to the training of models and their parameters. Section~\ref{sec:Results} presents the empirical findings of our investigations, followed by a careful analysis of the results. In Section~\ref{sec:Discussion}, we provide a critical discussion and contextualization within the scope of alternative approaches. Finally, Section~\ref{sec:Conclusion} synthesizes our findings and outlines potential directions for further research.

\section{Related Work}
\label{sec:related}

Fraud detection is a well-established area of research in the domain of AI. However, most methodologies focus on AI-generated NL content \cite{mitchell2023detectgpt, gehrmann2019gltr, solaiman2019release, islam2023distinguishing} rather than on code \cite{Hoq_Code_Detection, yang2023zeroshot}. Nevertheless, findings from text-based studies remain relevant, given the potential for cross-application and transferability of techniques. 

\subsection{Zero-shot detection}

One of the most successful models for differentiating between human- and AI-genera\-ted text is DetectGPT by \citeauthor{mitchell2023detectgpt}~\cite{mitchell2023detectgpt}, which employs zero-shot detection, i.e., it requires neither labeled training data nor specific model training. Instead, DetectGPT follows a simple hypothesis: Minor rewrites of model-generated text tend to have lower log probability under the model than the original sample, while minor rewrites of human-written text may have higher or lower log probability than the original sample. 
DetectGPT only requires log probabilities computed by the model of interest and random perturbations of the passage from another generic pre-trained LLM (e.g., \texttt{T5}~\cite{raffel2020exploring}). Applying this method to textual data of different origins, Mitchell et al.~\cite{mitchell2023detectgpt} report very good classification results, namely $\text{AUC} = 0.90 - 0.99$, which is considerably better than other zero-shot detection methods on the same data.

\citeauthor{yang2023zeroshot}~\cite{yang2023zeroshot} developed DetectGPT4Code, an adaptation of DetectGPT for code detection that also operates as a zero-shot detector, by introducing three modifications:
(\textbf{1}) Replacing \texttt{T5} with the code-specific \texttt{Incoder-6B} \cite{fried2023incoder} for code perturbations, addressing the need for maintaining code's syntactic and semantic integrity. (\textbf{2}) Employing smaller surrogate LLMs to approximate the probability distributions of closed, black-box models, like \texttt{GPT-3.5} \cite{brown2020language} or \texttt{GPT-4} \cite{openai2023gpt4}. (\textbf{3}) Using fewer tokens as anchors turned out to be better than the full-length code. Preliminary experiments found that the ending tokens are more deterministic given enough preceding text and thus better indicators. 
\citeauthor{yang2023zeroshot}~\cite{yang2023zeroshot} tested DetectGPT4Code on a relatively small set of $102$ Python and $165$ Java code snippets. Their results with $\text{AUC} = 0.70 - 0.80$ were clearly better than using plain DetectGPT ($\text{AUC}  = 0.50 - 0.60$) but still not reliable enough for practical use.

\subsection{Text Detectors Applied to Code} \label{subsec:RW_text_detectors}
Recently, several detectors like DetectGPT~\cite{mitchell2023detectgpt}, GPTZero~\cite{tian2023gptzero} and others~\cite{gehrmann2019gltr, solaiman2019release, islam2023distinguishing,guo2023RoBERTa-QA} were developed that are good at distinguishing AI-generated from human-generated NL texts, often with an accuracy better than $90\%$. This also makes it tempting to apply those NL text detectors to code snippets. As written above, \citeauthor{yang2023zeroshot}~\cite{yang2023zeroshot} used DetectGPT as a baseline for their code detection method.

There are two recent works ~\cite{wang2023evaluating,pan2024assessing} that compare a larger variety of NL text detectors on code detection tasks: \citeauthor{wang2023evaluating}~\cite{wang2023evaluating} collect a large code-related content dataset, among them $226$k code snippets, and apply $6$ different text detectors to it. When using the text detectors as-is, they only reach a low $\text{AUC} = 0.40 - 0.50$ on code snippets, which they consider unsuitable for reliable classification. In a second experiment, they fine-tune one of the open-source detectors (RoBERTa-QA~\cite{guo2023RoBERTa-QA}) by training it with a portion of their code data. It is unclear whether this fine-tuning and its evaluation used training and test samples originating from the same coding problem. Interestingly, after fine-tuning, they report a considerably higher $\text{AUC} = 0.77 - 0.98$. The authors conclude that ``While fine-tuning can improve performance, the generalization of the model still remains a challenge''.


\citeauthor{pan2024assessing}~\cite{pan2024assessing} provide a similar study on a medium-size database with $5$k code snippets, testing $5$ different text detectors on their ability to recognize the origin. As a special feature, they consider $13$ variant prompts.
They report an accuracy of $50-60\%$ for the tested detectors, only slightly better than random choice.

In general, text detectors work well on NL detection tasks but are not reliable enough on code detection tasks.

\subsection{Embedding- and Feature-Based Methods}
In modern LLMs, embeddings constitute an essential component, transforming text or code into continuous representations within a dense vector space of constant dimension, where proximity indicates similarity among elements. \citeauthor{Hoq_Code_Detection}~\cite{Hoq_Code_Detection} use a prior term frequency-inverse document frequency (\textbf{TF-IDF)}~\cite{salton1988term} embedding for classic ML algorithms, code2Vec~\cite{alon2018code2vec} and abstract syntax tree-based neural networks (\textbf{ASTNN})~\cite{zhang2019novel} for predicting the code's origin. While TF-IDF reflects the frequency of a token in a code snippet over a collection of code snippets, code2Vec converts a code snippet, represented as an abstract syntax tree (\textbf{AST}), into a set of path-contexts, linking pairs of terminal nodes. Subsequently, it computes the attention weights of the paths and uses them to compute the single aggregated weighted code vector. Similarly, ASTNN parses code snippets as an AST and uses preorder traversal to segment the AST into a sequence of statement trees, which are further encoded into vectors with pre-trained embedding parameters of Word2Vec~\cite{mikolov2013distributed}. These vectors are processed through a Bidirectional Gated Recurrent Unit (Bi-GRU)~\cite{tang2015document} to model statement naturalness, with pooling of Bi-GRU hidden states to represent the code fragment. \citeauthor{Hoq_Code_Detection}~\cite{Hoq_Code_Detection} used $3.162 \times 10^3$ human- and $3 \times 10^3$ ChatGPT-generated code snippets in Java from a CS1 course with a total of $10$ distinct problems, yielding $300$ solutions per problem. Further, they select $4 \times 10^3$ random code snippets for training and distribute the remaining samples equally on the test- and validation set. All models achieve similar accuracies, ranging from $0.90 - 0.95$. However, the small number of unique problems, the large number of similar solutions, and their splitting procedure render the results challenging to generalize beyond the study's specific context, potentially limiting the applicability of the finding to broader scenarios. 

\citeauthor{li2023discriminating}~\cite{li2023discriminating} present an interesting work where they generate features in three groups (lexical, structural layout, and semantic) for discrimination of code generated by ChatGPT from human-generated code. Based on these rich feature sets, they reach detection accuracies between $0.93 - 0.97$ with traditional ML classifiers like random forests (\textbf{RF}) or sequential minimal optimization (SMO).
Limitations of the method, as mentioned by the authors, are the relatively small ChatGPT code dataset ($1206$ code snippets) and the lack of prompt engineering 
(specific prompt instructions may lead to different results).

\section{Algorithms and Methodology}
\label{sec:methodology}
In this section, we present fundamental algorithms and describe the approaches mandatory for our methodology. We start by detailing the general prerequisites and preprocessing needed for algorithm application. Subsequently, we explicate our strategies for code detection and briefly delineate the models used for code sample classification.


\subsection{General Prerequisites} \label{sec:general_prerequisites}
Detecting fraudulent use of ChatGPT in software development or coding assessment scenarios requires an appropriate dataset.
Coding tasks consisting of a requirement text, human solutions, and test cases are of interest. To the best of our knowledge, one of the most used methods for fraudulent usage is representing the requirement text as the prompt for ChatGPT's input and using the output's extracted code for submission. Therefore, tasks that fulfill the above criteria are sampled from several coding websites, and human solutions are used as a baseline to compare to the code from ChatGPT. The attached test cases were applied to both the human and AI solutions before comparison to guarantee that the code is not arbitrary but functional and correct.
For generating code \texttt{gpt-3.5-turbo}~\cite{openai_api} was used, the most commonly used AI tool for fraudulent content, which also powers the application ChatGPT.

\subsection{ChatGPT}
Fundamentally, ChatGPT is a fine-tuned sequence-to-sequence learning~\cite{sutskever2014sequence} model with an encoder-decoder structure based on a pre-trained transformer~\cite{vaswani2017_NeurIPS, brown2020language}. Due to its positional encoding and self-attention mechanism, it can process data in parallel rather than sequentially, unlike previously used models such as recurrent neural networks~\cite{medsker2001recurrent} or long-short-term memory (\textbf{LSTM}) models~\cite{hochreiter1997long}. Just limited in the maximum capacity of input tokens, it is capable of capturing long-term dependencies. During inference, the decoder is detached from the encoder and is used solely to output further tokens. Interacting with the model requires the user to provide input that is then processed and passed into the decoder, which generates the output sequence token-by-token.
Once a token is generated, the model incorporates this new token into the input from the preceding forward pass, continuously generating subsequent tokens until a termination criterion is met. Upon completion, the model stands by for the next user input, seamlessly integrating it with the ongoing conversation. This process effectively simulates an interactive chat with a GPT model, maintaining the flow of the conversation.

\subsection{Embeddings}
Leveraging the contextual representation of embeddings in a continuous and constant space allows ML models to perform mathematical operations and understand patterns or similarities in the data. In our context, we use the following three models to embed all code snippets:

\begin{itemize}
    \item[] \textbf{TF-IDF} \cite{robertson2004understanding} 
    incorporates an initial step of prior tokenization of code snippets, setting the foundation to
    capture two primary components: (\textbf{1}) term frequency (TF), which is the number of times a token appears in a code snippet, and (\textbf{2}) inverse document frequency (IDF), which reduces the weight of tokens that are common across multiple code snippets. Formally, TF-IDF is defined as:
    \begin{align*}
        \text{TF-IDF}&(t,d) = \text{TF}(t,d) \cdot \text{IDF}(t), \\
        &\text{with TF}(t,d) = \frac{N_{t,d}}{N_{d}} \text{ and IDF}(t) = \log \left ( \frac{N}{N_t} \right ),
    \end{align*}
    where $N$ is the number of code snippets, $N_{t,d}$ the number of times token $t$ appears in code snippet $d$, $N_{d}$ the number of tokens in code snippet $d$ and $N_t$ the number of code snippets that include token $t$. The score emphasizes tokens that occur frequently in a particular code snippet but are less frequent in the entire collection of code snippets, thereby underlining the unique relevance of those tokens to that particular code snippet.
    \item[] \textbf{Word2Vec} \cite{mikolov2013efficient} is a neural network-based technique used to generate dense vector representations of words in a continuous vector space. It fundamentally operates on one of two architectures: (\textbf{1}) Skip-gram (\textbf{SG}), where the model predicts the surrounding context given a word, or (\textbf{2}) continuous bag of words (\textbf{CBOW}), where the model aims to predict a target word from its surrounding context. Given a sequence of words $w_1, \dots, w_T$, their objective is to maximize the average log probability:
    \begin{align*}
        &\frac{1}{T} \sum_{t=1}^{T}\sum_{\substack{-c\leq j \leq c \\ j \neq 0}}\begin{cases}
             \log \left (p \left (w_t | w_{t+j} \right ) \right ) & \text{ for SG}\\
             \log \left (p \left (w_{t+j} | w_t \right ) \right ) & \text{ for CBOW}
        \end{cases} \\
        & \text{with } p \left (w_{O} | w_{I} \right ) = \frac{\exp \left (v{'}_{w_{O}}^{\intercal} v_{w_{I}} \right )}{\sum_{w=1}^{W} \exp \left (v{'}_{w}^{\intercal} v_{w_{I}} \right )},
    \end{align*}
    where $v_w$ and $v{'}_{w}$ denote the input and output vector representation of word $w_i \in \mathcal{V}$ in the sequence of all words in the vocabulary, $W \in \mathbb{N}$ the number of words in that vocabulary $\mathcal{V}$, and $c \in \mathbb{N}$ the size of the training context. The probability of a word given its context is calculated by the softmax function with $ p \left (w_{O} | w_{I} \right )$. Training the model efficiently involves the use of hierarchical softmax and negative sampling to avoid the computational challenges of the softmax over large vocabularies~\cite{mikolov2013distributed}. 
    \item[] \textbf{OpenAI \textsc{Ada}}~\cite{openai_api} does not have an official paper, but we strongly suspect that a methodology related to OpenAI's paper~\cite{neelakantan2022text} was used to train the model. In their approach, \citeauthor{neelakantan2022text}~\cite{neelakantan2022text} use a contrastive objective on semantically similar paired samples $\{(x_i, y_i)\}_{i=1}^{N}$ and in-batch negative in training. Therefore, a transformers pre-trained encoder $E$ \cite{vaswani2017_NeurIPS}, initialized with GPT models~\cite{brown2020language, chen2020simple}, was used to map each pairs elements to their embeddings, and calculate the cosine similarity:
    \noindent
    \begin{minipage}{0.4\textwidth}
        \begin{align*}
            v_x &= E \left ( \left [\texttt{SOS} \right ]_x \oplus x \oplus \left [\texttt{EOS} \right ]_x \right ) \\
            v_y & = E \left ( \left [\texttt{SOS} \right ]_y \oplus y \oplus \left [\texttt{EOS} \right ]_y \right ) \\    
        \end{align*}
    \end{minipage} 
    and \hfill
    \begin{minipage}{0.47\textwidth}
        \begin{equation} \label{eq:cos_sim}
            \text{sim}(x,y) = \frac{v_x \cdot v_y}{\lVert v_x \rVert \cdot \lVert v_y \rVert},
        \end{equation}
    \end{minipage}
    where $\oplus$ denotes the operation of string concatenation and $\texttt{EOS}, \texttt{SOS}$ special tokens, delimiting the sequences. Fine-tuning the model includes contrasting the paired samples against in-batch negatives, given by supervised training data like natural language inference (\textbf{NLI})~\cite{NLI}. Mini-batches of $M$ samples are considered for training, which consist of $M-1$ negative samples from NLI, and one positive example $(x_i, y_i)$. Thus, the logits for one batch is a $M \times M$ matrix, where each logit is defined as $\hat{y} = \text{sim}\left (x_i,y_i \right ) \cdot \exp \left( \tau \right )$, where $\tau$ is a trainable temperature parameter. The loss is calculated as the cross entropy losses across each row and column direction, where positives examples lie on the diagonal of the matrix. Currently, embeddings from \textsc{Ada} can be obtained by using OpenAIs API, namely \texttt{text-embedding-ada-002}~\cite{openai_api}, which returns a non-variable dimension $\Tilde{x} \in \mathbb{R}^{1536}$.
\end{itemize}


\subsection{Supervised Learning Methods}
\label{subsec:SL_methods}
Feature extraction and embedding derivation constitute integral components in distinguishing between AI-generated and human-generated code, serving as inputs for classification models. Subsequently, we list the supervised learning (\textbf{SL}) models employed in our analysis: 

\begin{itemize}
    \item[] \textbf{Logistic Regression (LR)} \cite{logisticregression} which makes a linear regression model usable for classification tasks.
    
    \item[] \textbf{Classification and Regression Tree (CART)} \cite{cart} is a well-known form of decision trees (\textbf{DT}s) that offers transparent decision-making. Its simplicity, consisting of simple rules, makes it easy to use and understand.  

    \item[] \textbf{Oblique Predictive Clustering Tree (OPCT)} \cite{opct}: In contrast to regular DTs like CART, an OPCT split at a decision node is not restricted to a single feature, but rather a linear combination of features, cutting the feature space along arbitrary slanted (oblique) hyperplanes.
    
    \item[] \textbf{Random Forest (RF)} \cite{randomforest}: A random forest is an ensemble method, i.e., the application of several DTs, and is subject to the idea of bagging. RF tend to be much more accurate than individual DTs due to their ensemble nature, usually at the price of reduced interpretability.
    

    \item[] \textbf{eXtreme Gradient Boosting (XGB) \cite{xgboost}}: Boosting is an ensemble technique that aims to create a strong classifier from several weak classifiers. In contrast to RF with its independent trees, in boosting the weak learners are trained sequentially, with each new learner attempting to correct the errors of their predecessors. In addition to gradient boosting~\cite{friedman2001greedy}, XGB employs a more sophisticated objective function with regularization to prevent overfitting and improve computational efficiency. 
    
    \item[] \textbf{Deep Neural Network (DNN)} \cite{hopfield1982neural, dnn2}: A feedforward neural network with multiple layers. DNNs can learn highly complex patterns and hierarchical representations, making them extremely powerful for various tasks. However, they require large amounts of data and computational resources for training and their highly non-linear nature makes them, in contrast to other methods, somewhat of a ``black-box'', making it difficult to interpret their predictions.
\end{itemize}

\subsection{Gaussian Mixture Models}
\label{sec:gmm_methods}
Beyond SL methods, we also incorporate Gaussian mixture models (\textbf{GMM}s). Generally, a GMM is characterized by a set of $K$ 
Gaussian distributions $\mathcal{N}(x \vert \mu, \sigma)$. Each distribution $k = 1, \dots, K$ has a mean vector $\vec{\mu}_k$ and a covariance matrix $\Sigma_k$. Additionally, there are mixing coefficients $\psi_k$ associated with each Gaussian component $k$ 
, satisfying the condition $\sum_{k=1}^{K} \psi_k = 1$ to ensure the probability is normalized to $1$. Further, all components $k$ are initialized with $k$-means, modeling each cluster with the corresponding Gaussian distribution. The probability density function of a GMM is defined as:
\begin{equation*}
    p(\vec{x}) = \sum_{k=1}^{K} \psi_k \mathcal{N} \left ( \vec{x} | \vec{\mu}_k, \Sigma_k \right ).
\end{equation*}
The pre-defined clusters serve as starting point for optimizing the GMM with the expectation-maximization (\textbf{EM}) algorithm, which refines the model through iterative expectation and maximization steps. In the expectation step, it calculates the posterior probabilities $\hat{\gamma}_{ik}$ of data points belonging to each Gaussian component $k$, using the current parameter estimates according to Eq.~\eqref{eq:gmm_em_e}. Subsequently, the maximization step in Eq.~\eqref{eq:gmm_em_m} updates the model parameters ($\hat{\psi}_k$, $\hat{\vec{\mu}}_k$, $\hat{\Sigma}_k$) to maximize the data likelihood:

\noindent
\begin{minipage}{0.38\textwidth}
    \centering
    \begin{equation}
    \label{eq:gmm_em_e}
    \hat{\gamma}_{ik} = \frac{\hat{\psi}_k \mathcal{N}(\vec{x}_i | \hat{\vec{\mu}}_k, \hat{\Sigma}_k)}{\sum_{j=1}^{K} \hat{\psi}_j \mathcal{N}(\vec{x}_i | \hat{\vec{\mu}}_j, \hat{\Sigma}_j)}
    \end{equation}    
\end{minipage} \hfill
\begin{minipage}{0.6\textwidth}
    \centering
    \begin{equation}
    \label{eq:gmm_em_m}
        \begin{split}
            \hat{\psi}_k &= \frac{1}{N} \sum_{i=1}^{N} \hat{\gamma}_{ik}, \quad \hat{\vec{\mu}}_k = \frac{1}{N \hat{\psi}_k} \sum_{i=1}^{N} \hat{\gamma}_{ik} \vec{x}_i \\
    \hat{\Sigma}_k &= \frac{1}{(N-1) \hat{\psi}_k} \sum_{i=1}^{N} \hat{\gamma}_{ik} (\vec{x}_i - \hat{\vec{\mu}}_k)(\vec{x}_i - \hat{\vec{\mu}}_k)^{\intercal}
        \end{split}
    \end{equation}
\end{minipage}

The iterative repetition of this process guarantees that at least one local optimum and possibly the global optimum is always achieved.

\section{Experimental Setup}
\label{sec:exp_setup}
In this section, we outline the requirements to carry out our experiments. We cover basic hard- and software components, as well as the collection and preprocessing of data to apply the methodology and models presented in the previous section.

For data preparation and all our experiments, we used Python version $3.10$ with different packages, as delineated in our repository \url{https://github.com/MarcOedingen/ChatGPT-Code-Detection} (accessed on \today). Due to large amounts of code snippets, we recommend a minimum of $32$ GB of RAM, especially when experimenting with Word2Vec.


\subsection{Data Collection}
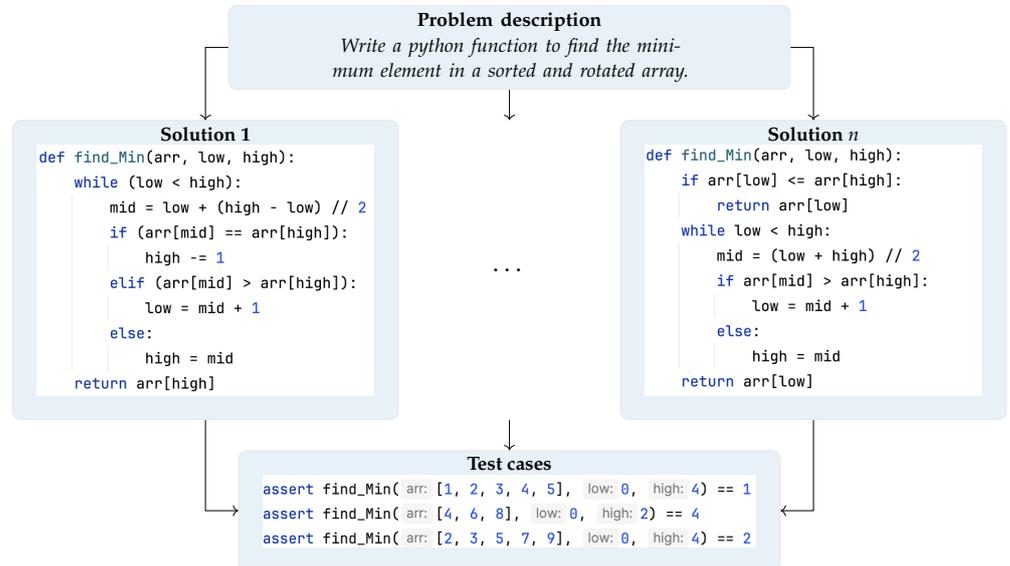
\begin{figure}[h]
    \centering
    \scalebox{0.8}{
    \begin{tikzpicture}[node distance=1cm and 1cm, auto]
        \definecolor{color_def}{RGB}{20,118,183}
        \node (problem_description) [rectangle, rounded corners, minimum width=8cm, minimum height=1cm, draw=black, align=center, fill=color_def, opacity=0.1, text opacity=1, text width=9cm] {\small \textbf{Problem description} \\ \small \textit{Write a python function to find the minimum element in a sorted and rotated array.}};
        
        \node[below=0.5cm of problem_description, xshift=-5cm, rectangle, rounded corners, draw=black, inner sep=4mm, align=center, fill=color_def, opacity=0.1, text opacity=1, label={[anchor=north,inner sep=1mm]north:\small \textbf{Solution 1}}] (solution_box1) {\includegraphics[width=0.4\linewidth]{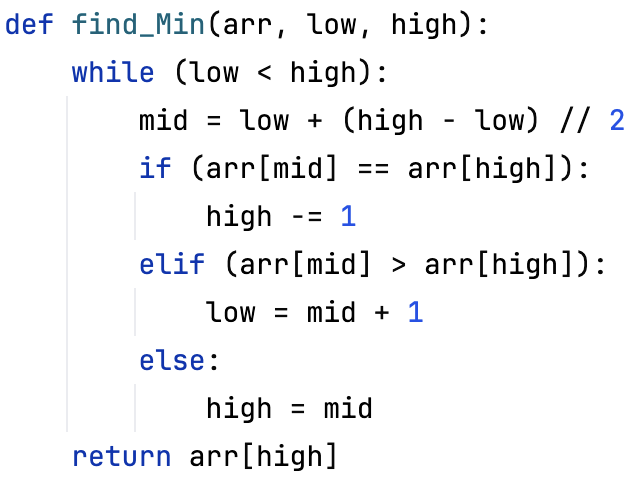}};
        
        \node[below=0.5cm of problem_description, xshift=5cm, rectangle, rounded corners, draw=black, inner sep=4mm, align=center, fill=color_def, opacity=0.1, text opacity=1, label={[anchor=north,inner sep=1mm]north:\small \textbf{Solution $n$}}] (solution_box2) {\includegraphics[width=0.4\linewidth]{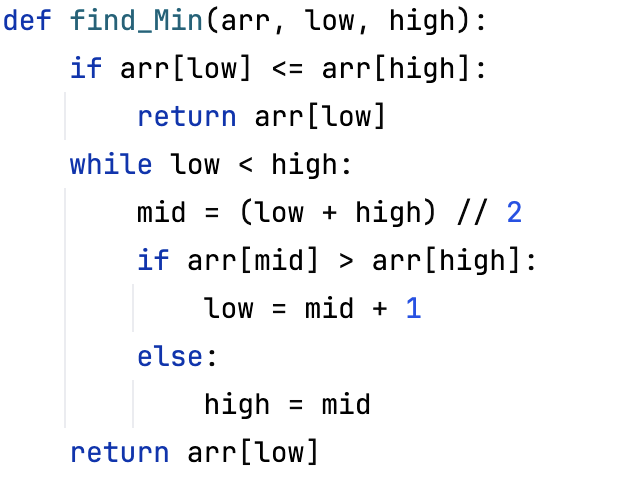}};

        \node[below=0.5cm of problem_description] (begin_solutions) {};

        \node at ($(solution_box1)!0.5!(solution_box2)$) (dots) {\large $\dots$};
        
        \node[below=of solution_box1, below= 0.5cm of solution_box2, xshift=-5cm, rectangle, rounded corners, draw=black, inner sep=4mm, align=center, fill=color_def, opacity=0.1, text opacity=1, label={[anchor=north,inner sep=1mm]north:\small \textbf{Test cases}}] (test_cases) {\includegraphics[width=0.585\linewidth]{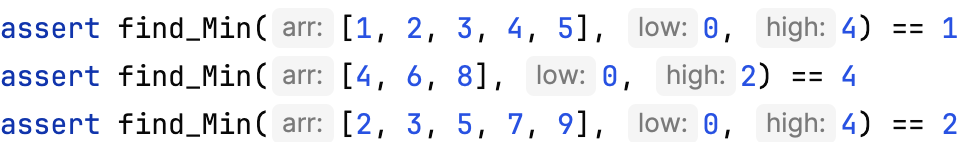}};

        \node[above=0.5cm of test_cases] (end_solutions) {};
        
        \draw[->] (problem_description.west) -| (solution_box1);
        \draw[->] (problem_description.east) -| (solution_box2);
        \draw[->] (problem_description.south) -- (begin_solutions);
        \draw[->] (end_solutions.south) -- (test_cases.north);
        \draw[->] (solution_box1.south) |- (test_cases.west);
        \draw[->] (solution_box2.south) |- (test_cases.east);
        
    \end{tikzpicture}}
    \caption{Example of row in dataset}
    \label{fig:example_sample}
\end{figure}

As previously delineated in the general prerequisites, an ideal dataset for our intended purposes is characterized by the inclusion of three fundamental elements: (\textbf{1}) Problem description, (\textbf{2}) one or more human solutions, and (\textbf{3}) various test cases. This is exemplified in Figure~\ref{fig:example_sample}. The problem description (1) should clearly contain the minimum information required to solve a programming task or to generate a solution through ChatGPT. In contrast, unclear problem descriptions may lead to solutions that overlook the main problem, thereby lowering the solution quality and potentially omitting useful solutions from the limited available samples. The attached human solutions (2) for a coding problem play an important role in the subsequent analysis and serve as referential benchmark for the output of ChatGPT. Furthermore, a set of test cases (3) facilitates the elimination of syntactically correct solutions that do not fulfill the functional requirements specified in the problem description. To this end, a controlled environment is created in which the code's functionality is rigorously tested, preventing the inclusion of snippets of code that are based on incorrect logic or could potentially produce erroneous output. Hence, we strongly focus on syntactic and executable code but ignore a possibly typical behavior of ChatGPT in case of uncertainties or wrong answers. For some problems, a function is expected to solve them, while others expect a console output. We have considered both by using either the function name or the entire script, referred to as the entry point, for the enclosed test cases.
\noindent

Programming tasks from programming competitions are particularly suitable for the above criteria; see Table~\ref{tab:total_samples} for the sources of the coding problems. Due to the variety and high availability of such tasks in Python, we decided to use this programming language. Thereby, we exclusively include human solutions from a period preceding the launch of ChatGPT. We used OpenAIs \texttt{gpt3.5-turbo} API with the default parameters to generate code.

The OpenAI report~\cite{openai2023gpt4} claims that \texttt{GPT-3.5} has an accuracy of $48.1\%$ in a zero-shot 
evaluation for generating a correct solution on the HumanEval dataset~\cite{hed}. After a single generation, our experimental verification yielded a notably lower average probability of $21.3\%$. 
Due to the low success rate, we conducted five distinct API calls for each collected problem. This strategy improved the accuracy rate considerably to $45.6\%$, converging towards the accuracy reported in~\cite{openai2023gpt4} and substantiating the theory that an increment in generation attempts correlates positively with heightened accuracy levels~\cite{hed}. Furthermore, it is noteworthy that the existence of multiple AI-generated solutions for a single problem does not pose an issue, given that the majority of problems possess various human solutions; see Table~\ref{tab:total_samples} in column `before pre-processing' for the number of problems $n_{\textsc{problems}}$, average human solutions per problem $\bar{n}_{\textsc{problems}}$ and the total human samples $n_{\textsc{samples}}$.

When generating code with \texttt{gpt3.5-turbo}, the prompt strongly influences success rate. Prompt engineering is a separate area of research that aims to utilize the intrinsic capabilities of an LLM while mitigating potential pitfalls related to unclear problem descriptions or inherent biases. During the project, we tried different prompts to increase the yield of successful 
solutions. Our most successful prompt, which we subsequently used, is the following: `\textit{Question:} $<$Coding\_Task\_Description$>$ \textit{Please provide your answer as Python code.} \textit{Answer:}''. Other prompts, i.e., ``\textit{Question:} $<$Coding\_Task\_Description$>$ \textit{You are a developer writing Python code. Put all python code \$PYTHON in between [[[\$PYTHON]]]. Answer:}'', led to a detailed explanation of the problem and an associated solution strategy of the model, but without the solution in code.

\subsection{Data Preprocessing}
\label{sec:data_preproc}
Based on impurities in both the GPT-generated and human solutions, the data must be preprocessed before it can be used as input for ML models. Hence, we first extracted the code from the GPT-generated responses and checked whether it and the human solutions can be executed, 
reducing the whole dataset to $3.68 \times 10^{5}$ samples. 
This also eliminated missing values due to miscommunication with the API, server overloads, or the absence of Python code in the answer. Further, to prevent the overpopulation of particular code snippet subsets, we removed duplicates in both classes. 
A duplicate is a code snippet for problem $P$ that is identical to another code snippet for the same problem $P$.
This first preprocessing step leaves us with $3.14 \times 10^{5}$ samples. 

Numerically, the largest collapse for the remaining samples, and especially for the GPT-generated code snippets, is given by the application of the test cases. This reduces the number of remaining GPT samples by $72.39\%$ and the number of human samples by $28.59\%$, leaving a total of $1.71 \times 10^{5}$ samples. Furthermore, we consider a balanced dataset so that our models are less likely to develop biases or favor a particular class, reducing the risk of overfitting and making the evaluation of the model's performance more straightforward. 
Given $n$ individual coding problems $P_i$, 
$i=1, \dots, n$, with $h_i$ 
human solutions and $g_i$ 
GPT solutions, we take the minimum 
$k_i = \min(h_i, g_i)$ and choose $k_i$ random and distinct solutions from each of the two classes for $P_i$. 
The figures for human samples after the pre-processing procedure are listed in Table~\ref{tab:total_samples} in column `after pre-processing'. Based on a balanced dataset, there are as many average GPT solutions $\bar{n}_\textsc{solutions}$ and total GPT samples $n_\textsc{samples}$ as human solutions and samples for each source in the final processed dataset. 
Thus, the pre-processed, balanced and cleaned dataset contains $3.14 \times 10^{4}$ samples in total.

\begin{table}[h]
    \caption{Code datasets overview: $n_{\textsc{problems}}$ - number of distinct problems, $\bar{n}_{\textsc{solutions}}$ - average number of human solutions per problem, and $n_{\textsc{samples}}$ - total human samples per data source. 
    }
    \label{tab:total_samples}
    \centering
    \small
    \begin{tabular}{llll|lll}
    & \multicolumn{3}{c}{before pre-processing} & \multicolumn{3}{c}{after pre-processing} \\
    \cmidrule(lr){2-4} \cmidrule(lr){5-7}
    Dataset   &   $n_{\textsc{problems}}$  &   $\bar{n}_{\textsc{solutions}}$  & $n_{\textsc{samples}}$ &   $n_{\textsc{problems}}$  &   $\bar{n}_{\textsc{solutions}}$  & $n_{\textsc{samples}}$ \\
    \toprule
    APPS \cite{hendrycks2021measuring}  &   $1.00 \times 10^4$  &   $21$    &   $2.10 \times 10^5$  & $1.95 \times 10^3$    &    $1.1$ &   $2.17 \times 10^3$ \\
    CCF \cite{codechef}    &   $1.56 \times 10^3$  &   $18$    &   $2.81 \times 10^4$   & $4.89 \times 10^{2}$  &   $2.4$   &   $1.17 \times 10^{3}$    \\
    CC \cite{codecontest}     &   $8.26 \times 10^3$  &   $15$    &   $1.23 \times 10^5$  & $3.11 \times 10^{3}$    &   $3.0$ &   $9.30 \times 10^{3}$    \\
    HAEA \cite{hackerearth}   &   $1.49 \times 10^3$  &   $16$    &   $2.38 \times 10^4$  & $5.24 \times 10^{2}$    &   $3.9$ &   $2.05 \times 10^{3}$    \\
    HED \cite{hed}     &   $1.64 \times 10^2$  &   $1$     &   $1.64 \times 10^2$  & $1.27 \times 10^{2}$   &   $1.0$ &   $1.27 \times 10^{2}$    \\
    MBPPD \cite{MBPPD}  &   $9.74 \times 10^2$  &   $1$     &   $9.74 \times 10^2$  & $6.91 \times 10^{2}$  &   $1.2$ &   $8.40 \times 10^{2}$    \\
    MTrajK \cite{mtrajk} &   $1.44 \times 10^2$  &   $1$     &   $1.44 \times 10^2$  & $3.20 \times 10^{1}$ &   $1.0$ &   $3.20 \times 10^{1}$    \\
    \midrule
    \textbf{Sum (Average)}   &   $\mathbf{2.26 \times 10^4}$   &  $\mathbf{(10)}$   &   $\mathbf{3.86 \times 10^5}$ &   $\mathbf{7.01 \times 10^{3}}$    &  $\mathbf{(1.9)}$    &   $\mathbf{1.57 \times 10^{4}}$ \\
    \bottomrule
    \end{tabular}
\end{table}

\subsection{Optional Code Formatting}
\begin{figure}[h]
    \centering
    \includegraphics[width=0.95\textwidth]{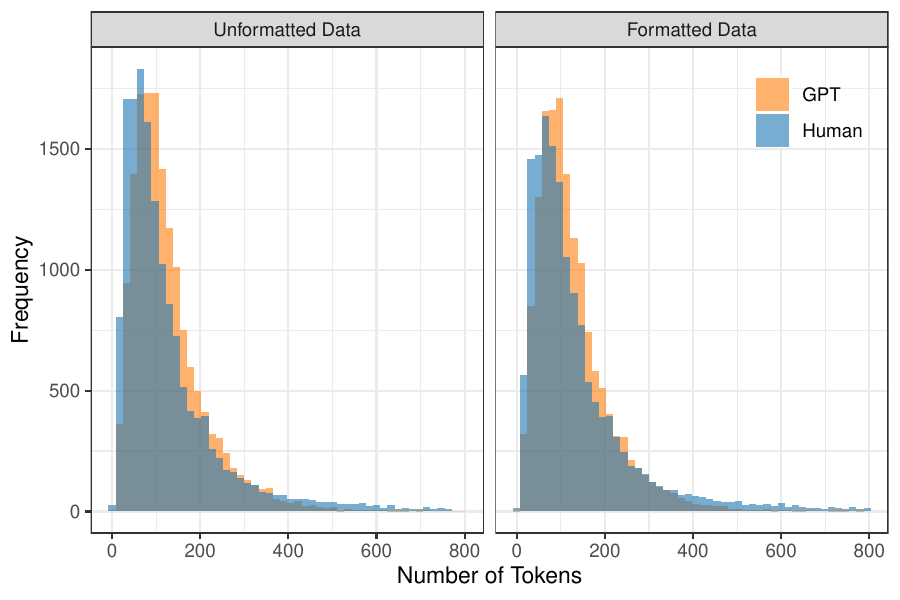}
    \caption{Distribution of the code length (number of tokens according to \texttt{cl100k\_base} encoding) across the unformatted and formatted dataset. Values larger than the $99\%$ quantile were removed to avoid a distorted picture.}
    \label{fig:code_length}
\end{figure}

A discernible method for distinguishing between the code sources lies in the analysis of code formatting patterns. Variations in these patterns may manifest through the presence of spaces over tabs for indention purposes or the uniform application of extended line lengths. Thus, we use the Black code formatter~\cite{black}, a Python code formatting tool, for both human- and GPT-generated code, standardizing all samples into a uniform formatting style in an automated way. This methodology effectively mitigates the model's tendency to focus on stylistic properties of the code. Consequently, it allows the models to emphasize more significant features beyond mere formatting. A comparison of the number of tokens for all code snippets of the unformatted and formatted datasets is shown in Figure~\ref{fig:code_length}.



\subsection{Training / Test Set Separation}
\label{sec:train_test}
In dividing our dataset into training and test sets, we employed a problem-wise division, allocating $80\%$ of the problems to the training set and $20\%$ to the test set instead of a sample-wise approach. This decision stems from our dataset's structure, which includes multiple solutions per problem. The sample-wise approach could include similar solutions for the same problem within the training and test sets. We opted for a problem-wise split to avoid this issue and enhance the model's generalization, ensuring the model is tested on unseen problem instances. Additionally, we repeat each experiment ten times for statistical reliability, each time using another seed for a different distribution of problems into training and test sets.

\subsection{Modeling Parameters and Tokenization}
\label{sec:model_params}
For all SL methods, we use the default parameters proposed by scikit-learn~\cite{scikit-learn} for RF, GB, LR, and DT, those of xgboost~\cite{xgboost} for XGB, those of spyct~\cite{opct} for OPCT, and those of TensorFlow~\cite{tensorflow2015-whitepaper} for DNN with two notable exceptions for DNNs (\textbf{1}) and OPCTs (\textbf{2}). For DNNs (1), we configure the network architecture to \texttt{[1536, 768, 512, 128, 32, 8, 1]}, employing the \texttt{relu} activation function across all layers except for the output layer, where \texttt{sigmoid} was used alongside binary cross-entropy as the loss function. In response to the strong fluctuation of the OPCTs (2), we create $10$ individual trees and select the best of them. Additionally, we standardized the number of Gaussian components $K=2$ for all experiments with GMMs.

While embedding \textsc{Ada} uses internal tokenization which is the \texttt{cl100k\_base} encoding,
we must explicitly tokenize the prepared formatted or unformatted code for TF-IDF and Word2Vec. We decide to use the same tokenization \texttt{cl100k\_base} encoding implemented in the \texttt{tiktoken} library~\cite{openai_api} which was uniformly applied across all code snippets. 
Given the fixed size of the embedding of \texttt{text-embedding-ada-002} with $\Tilde{x} \in \mathbb{R}^{1536}$, we adopted this dimensionality for the other embeddings. For TF-IDF, we retain sklearn's default parameters, while for gensim's~\cite{rehurek2011gensim} 
Word2Vec, we adjusted the threshold for a word's occurrence in the vocabulary to $\texttt{min\_count}=1$ and use CBOW as the training algorithm. 

\section{Results}
\label{sec:Results}

In this section, we present the primary outcomes from deploying the models introduced in Section~\ref{sec:methodology}, operating on the pre-processed dataset as shown in Section~\ref{sec:exp_setup}.
We discuss the impact of different kinds of features (human-designed vs. embeddings) and the calibration of ML models. We then put the results into perspective by comparing them to the performance of untrained humans and a Bayes classifier.

\subsection{Similarities between Code Snippets}
\label{sec:similarities}
The representation of the code snippets as embeddings describes a context-rich and high-dimensional vector space. However, the degree of similarity among code snippets within this space remains to be determined. Based on our balanced dataset and the code's functionality, we assume that the code samples are very similar. 
They are potentially even more similar when a code formatter, e.g., the Black code formatter~\cite{black}, is used, which presents the models with considerable challenges in distinguishing subtle differences.

\begin{figure}[h]
\centering
\includegraphics[width=0.95\textwidth]{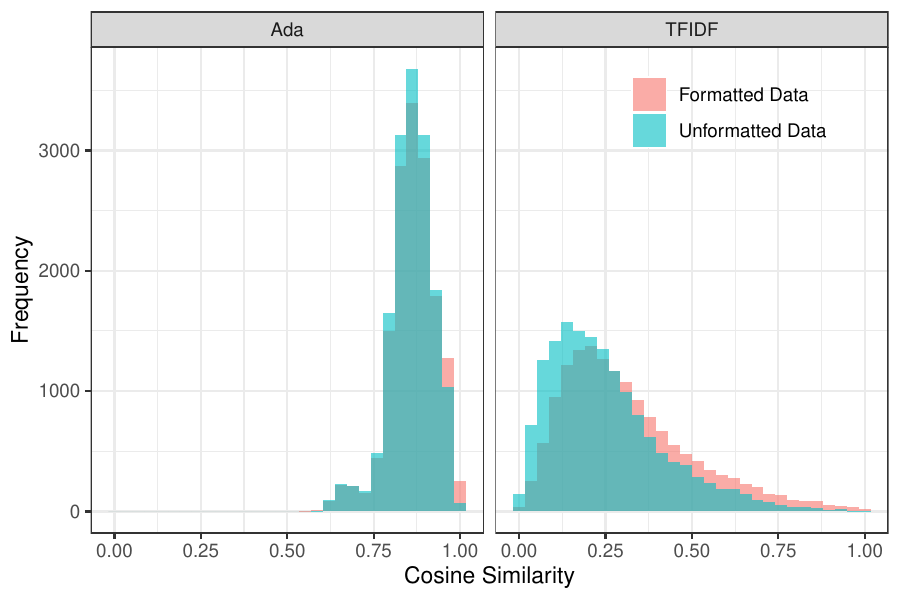}
\caption{Cosine similarity between all human and GPT code samples embedded using \textsc{Ada} and \textsc{TFIDF}, both formatted and unformatted. 
}
\label{fig:cosine_sim_hist}
\end{figure}

Mathematically, similarities in high-dimensional spaces can be particularly well calculated using cosine similarity. Let $H_P, G_P \in \mathcal{C}$ be code snippets for problem $P$ originating from humans and GPT, respectively. We compute the cosine similarity equivalent to Eq.~\eqref{eq:cos_sim} as sim$(H_P, G_P) = \frac{H_P \cdot G_P}{ \lVert H_P \rVert \lVert G_P \rVert}$.
Concerning the embeddings of all formatted and unformatted code snippets generated by \textsc{Ada}, the resulting distributions of cosine similarities are presented in Figure~\ref{fig:cosine_sim_hist}. This allows us to mathematically confirm our assumption that the embeddings of the codes are very similar. 
The cosine similarities for both the embeddings of the formatted and unformatted code samples in the \textsc{Ada}-case are approximately normally distributed, resulting in very similar mean and standard deviation: $\bar{x}_{\textsc{form}} = 0.859 \pm 0.065$ and $\bar{x}_{\textsc{unform}} =0.863 \pm 0.067$. Figure~\ref{fig:cosine_sim_hist} also shows the \textsc{TF-IDF} embeddings for both datasets. In contrast to the \textsc{Ada} embeddings, significantly lower cosine similarities can be identified. We suspect that this discrepancy arises because \textsc{TF-IDF} embeddings are sparse and based on exact word matches. In contrast, \textsc{Ada} embeddings are dense and capture semantic relationships and context. Finally, the average cosine similarities in the \textsc{TF-IDF}-case are $\bar{y}_{\textsc{form}} = 0.316 \pm 0.189$ for the formatted dataset and $\bar{y}_{\textsc{unform}} =0.252 \pm 0.166$ for the unformatted dataset.

Figure~\ref{fig:cosine_sim_hist} demonstrates that the cosine similarities of embeddings vary largely, depending on the kind of embedding.
But, as the results in Section~\ref{sec:embed_features} will show, ML models can effectively detect the code's origin from those embeddings. However, embeddings are black-box in the sense that the meaning of certain embedding dimensions is not explainable to humans.
Consequently, we also investigated 
features that can be interpreted by humans to avoid the black-box setting with embeddings.

\subsection{Human-designed Features (white-box)}
\label{sec:human_designed}
A possible and comprehensible differentiation of the code samples can be attributed to their formatting. 
Even if these differences are not immediately visible to the human eye, they can be determined with the help of calculations. 
To illustrate this, we have defined the features in Table~\ref{tab:exp_features}. We assess their applicability using the presented SL models from Section~\ref{subsec:SL_methods}. The results for the unformatted samples are displayed in Table~\ref{tab:evaluation_new_unformatted}, and for the formatted samples in Table~\ref{tab:evaluation_new_formatted}.


\begin{table}[h!]
\caption{Detailed description of all human-designed features.}
\label{tab:exp_features}
\begin{tabularx}{\columnwidth}{p{.325\columnwidth} p{.125\columnwidth} p{.47\columnwidth}}
\toprule
\small
Feature &   Abbreviation    &   Description\\ 
\midrule
Number of leading whitespaces  &   $n_{\textsc{lw}}$   &   The code sample is divided into individual lines, and the number of leading whitespaces is summed up.  \\
\midrule
Number of empty lines   &   $n_{\textsc{el}}$   &   The code sample is divided into individual lines, and the number of empty is summed up. \\
\midrule
Number of inline whitespaces    &   $n_{\textsc{iw}}$   &   The code sample is divided into individual lines, and the number of spaces within the trimmed content of each line is summed up. \\
\midrule
Number of punctuations  &   $n_{\textsc{pt}}$   &   The code sample is filtered using the regular expression {[}\textasciicircum{}\textbackslash{}w\textbackslash{}s{]} to isolate punctuation characters, which are then counted. \\
\midrule
Maximum line length &   $n_{\textsc{ml}}$   &   The code sample is split by every new line, and the maximum count of characters of a line is returned. \\
\midrule
Number of trailing whitespaces  &   $n_{\textsc{tw}}$   &   The code sample is divided into individual lines, and the number of trailing whitespaces is summed up. \\
\midrule
Number of lines with leading whitespaces    &   $n_{\textsc{lwl}}$  &   The code sample is split by every new line, and all lines starting with whitespaces are summed up. \\
\bottomrule
\end{tabularx}
\end{table}

\begin{figure}[h]
\includegraphics[width=\textwidth]{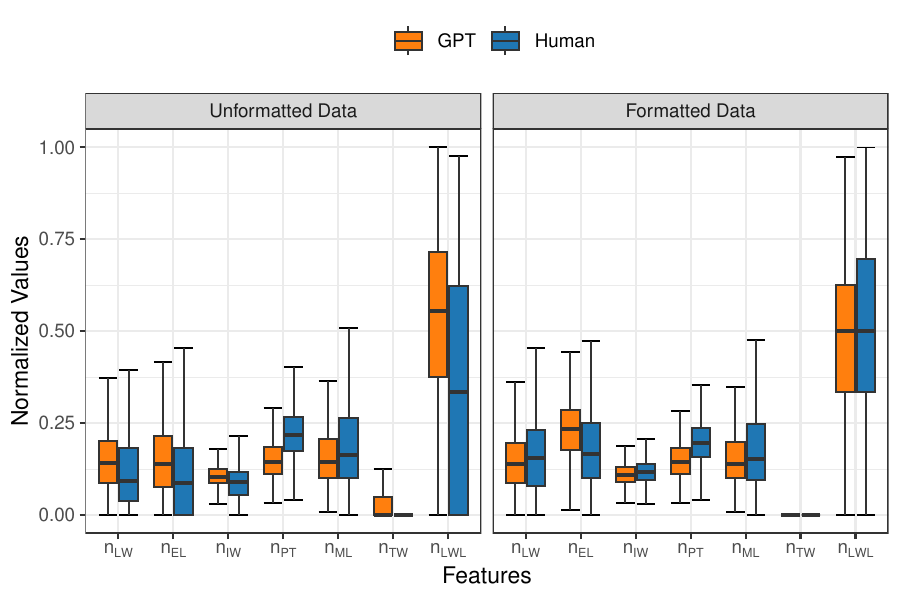}
\caption{Box plot of human-designed features for formatted and unformatted dataset.}
\label{fig:explainable_feature_boxplots}
\end{figure}

We find that the selected features capture a large proportion of the differences on the unformatted dataset. The XGB model stands out as the most effective one, achieving an average accuracy of $88.48\%$ across various problem splits. However, when assessing the formatted dataset, there is a noticeable performance drop, with the XGB model's effectiveness decreasing by approximately $8$ percentage points across all metrics. While outperforming all other models on the unformatted dataset, XGB slightly trails behind the RF model in the formatted dataset, where the RF model leads with an accuracy of $80.50\%$. This aligns with our expectations that the differences are minimized when employing formatting. While human-designed features are valuable for distinguishing unformatted code, their effectiveness diminishes significantly when formatting variations are reduced. This fact is further supported by Figure~\ref{fig:explainable_feature_boxplots}, which reflects the normalized values of the individual features from Table~\ref{tab:exp_features} over the two data sets. The figure demonstrates 
an alignment of the feature value distributions for humans and ChatGPT after formatting
, especially evident in features such as $n_\textsc{lwl}$ and $n_\textsc{tw}$, which show nearly identical values post-formatting.

Despite their astonishing performance at first glance, human-designed features do not lead to near-perfect classification results. This is likely due to the small number of features and the fact that they do not capture all available information. ML models that use the much higher-dimensional (richer) embedding space address these limitations by capturing implicit patterns not readily recognizable by human-designed features.

\subsection{Embedding Features (black-box)}
\label{sec:embed_features}
ML models operating on embeddings achieve superior performance compared to the models using the human-designed features; see Table~\ref{tab:evaluation_new_unformatted} and Table~\ref{tab:evaluation_new_formatted} for results on the unformatted and formatted dataset, respectively.
The results on the unformatted dataset show that the highest values are achieved by XGB + \textsc{TF-IDF} with an accuracy greater than $98\%$ and an astonishing AUC value of $99.84\%$. With a small gap to XBG + \textsc{TF-IDF} across all metrics, RF + \textsc{TF-IDF} is in second place. With less than one percentage point difference to XGB + \textsc{TF-IDF}, DNN + \textsc{Ada} is in third place. An identical ranking of the top 3 models can be found on the formatted dataset. As with the human-designed features, \textit{`formatted'} shows weaker performance than \textit{`unformatted'}, with a deterioration of about $4$ percentage points on all metrics except AUC, which only decreased by about $1$ percentage point. In a direct comparison of models based on either human-designed features or embeddings, the best-performing models of each type show a difference of approximately $10$ and $14$ percentage points across all metrics except the AUC score on the unformatted and formatted dataset, respectively. 

The disparities in performance between the white- and black-box approaches, despite employing identical models, highlight the significance of embedding techniques. While traditional feature engineering is based on domain-specific expertise or interpretability, it often cannot capture the complex details of code snippets as effectively as embedding features.

\begin{table*}[t]
\vspace{0.1cm}
\caption{Final results -- Shown are the mean $\mu\, \pm$ standard deviation $\sigma$ from $10$ independent runs on \textbf{unformatted} data; each run corresponds to a different seed, i.e., different distribution of training and test samples. \textbf{Boldface}: column maximum of $\mu$ in the `Human-designed (white-box)' box and for each individual embedding model in the `Embedding (black-box)' box.
}
\label{tab:evaluation_new_unformatted}
\centering
\small
\begin{tabular}{llllll}
\toprule
& \multicolumn{5}{c}{Unformatted} \\ \cmidrule(lr){2-6}
Model         & Accuracy ($\%$)    &   Precision ($\%$)   &   Recall ($\%$) &   F1 ($\%$)   &   AUC ($\%$)  \\
\midrule
& \multicolumn{5}{c}{Human-designed (white-box)} \\ 
\midrule
\textsc{CART}   &   $82.54 \pm 0.23$    &   $82.70 \pm 0.36$    &   $82.29 \pm 0.57$  &   $82.49 \pm 0.26$    &   $82.54 \pm 0.23$    \\
\textsc{DNN}    &   $85.37 \pm 0.80$    &   $84.58 \pm 2.73$    &   $86.84 \pm 4.77$  &   $85.54 \pm 1.21$    &   $93.73 \pm 0.42$    \\
\textsc{GMM}    &   $79.68 \pm 0.50$    &   $74.74 \pm 0.66$    &   $89.74 \pm 0.58$    &   $81.55 \pm 0.39$    &   $89.05 \pm 0.25$    \\
\textsc{LR}     &   $76.69 \pm 0.63$	&	$74.80 \pm 0.64$	&	$80.50 \pm 0.94$  &	  $77.54 \pm 0.63$    &	  $84.57 \pm 0.41$    \\
\textsc{OPCT}   &   $84.12 \pm 0.60$	&	$80.89 \pm 1.95$	&	$89.52 \pm 2.37$  &	  $84.94 \pm 0.43$    &	  $89.60 \pm 0.81$    \\
\textsc{RF}     &   $88.10 \pm 0.40$	&	$87.29 \pm 0.41$	&	$89.19 \pm 0.66$  &	  $88.23 \pm 0.41$    &	  $95.34 \pm 0.22$    \\
\textsc{XGB}    &   $\mathbf{88.48 \pm 0.26}$   &   $\mathbf{87.39 \pm 0.46}$   &   $\mathbf{89.93 \pm 0.42}$   &   $\mathbf{88.64 \pm 0.24}$   &   $\mathbf{95.59 \pm 0.20}$   \\
\midrule
& \multicolumn{5}{c}{Embedding (black-box)} \\ 
\midrule
& \multicolumn{5}{c}{\textsc{Ada}} \\ \cmidrule(lr){1-6}
\textsc{CART}   &   $80.82 \pm 0.50$    &   $80.63 \pm 0.55$    &   $81.46 \pm 0.67$    &   $79.83 \pm 1.03$    &   $80.82 \pm 0.50$    \\
\textsc{DNN}    &   $\mathbf{97.79 \pm 0.36}$    &   $\mathbf{97.40 \pm 0.77}$    &   $\mathbf{98.22 \pm 0.60}$    &   $\mathbf{97.80 \pm 0.35}$    &   $\mathbf{99.76 \pm 0.05}$ \\
\textsc{GMM}    &   $92.71 \pm 0.39$    &   $95.10 \pm 0.48$    &   $90.06 \pm 0.66$    &   $92.51 \pm 0.42$    &   $97.37 \pm 0.21$    \\
\textsc{LR}     &   $95.87 \pm 0.13$    &   $95.93 \pm 0.13$    &   $94.60 \pm 0.24$    &   $97.30 \pm 0.25$    &   $99.06 \pm 0.07$    \\
\textsc{OPCT}   &   $95.53 \pm 0.27$    &   $95.59 \pm 0.25$    &   $94.35 \pm 0.81$    &   $96.87 \pm 0.65$    &   $97.24 \pm 0.46$    \\
\textsc{RF}     &   $92.39 \pm 0.35$    &   $92.55 \pm 0.33$    &   $90.67 \pm 0.63$    &   $94.52 \pm 0.55$    &   $97.85 \pm 0.14$    \\
\textsc{XGB}    &   $95.05 \pm 0.23$    &   $95.09 \pm 0.22$    &   $94.30 \pm 0.42$    &   $95.89 \pm 0.30$    &   $98.94 \pm 0.09$    \\
\midrule
& \multicolumn{5}{c}{\textsc{TF-IDF}} \\ \cmidrule(lr){1-6}
\textsc{CART}   &   $95.82 \pm 0.31$    &   $95.82 \pm 0.31$    &   $95.93 \pm 0.26$    &   $95.71 \pm 0.52$    &   $95.82 \pm 0.31$    \\  
\textsc{DNN}    &   $97.61 \pm 0.30$    &   $97.79 \pm 0.79$    &   $97.45 \pm 1.17$    &   $97.61 \pm 0.32$    &   $99.68 \pm 0.05$    \\
\textsc{GMM}    &   $94.60 \pm 0.32$    &   $95.24 \pm 0.26$    &   $93.89 \pm 0.60$    &   $94.56 \pm 0.34$    &   $96.18 \pm 0.26$	\\
\textsc{LR}     &   $97.06 \pm 0.24$    &   $97.09 \pm 0.24$    &   $96.11 \pm 0.28$    &   $98.09 \pm 0.31$    &   $99.46 \pm 0.05$    \\
\textsc{OPCT}   &   $96.72 \pm 0.26$    &   $96.72 \pm 0.26$    &   $96.78 \pm 0.50$    &   $96.66 \pm 0.50$    &   $97.62 \pm 0.41$    \\  
\textsc{RF}     &   $98.04 \pm 0.11$    &   $\mathbf{98.04 \pm 0.11}$    &   $97.77 \pm 0.18$    &   $\mathbf{98.31 \pm 0.23}$    &   $99.72 \pm 0.05$    \\  
\textsc{XGB}    &   $\mathbf{98.28 \pm 0.09}$    &   $97.87 \pm 0.15$    &   $\mathbf{98.70 \pm 0.22}$    &   $98.28 \pm 0.09$    &   $\mathbf{99.84 \pm 0.02}$    \\
\midrule
& \multicolumn{5}{c}{\textsc{Word2Vec}} \\ \cmidrule(lr){1-6}
\textsc{GMM}    &   $93.57 \pm 0.20$    &   $92.49 \pm 0.37$    &   $94.86 \pm 0.31$    &   $93.66 \pm 0.19$    &   $94.14 \pm 0.14$    \\
\bottomrule
\end{tabular}
\end{table*}

\begin{figure}[tbp]
    \centering
    \includegraphics[width=0.8\textwidth]{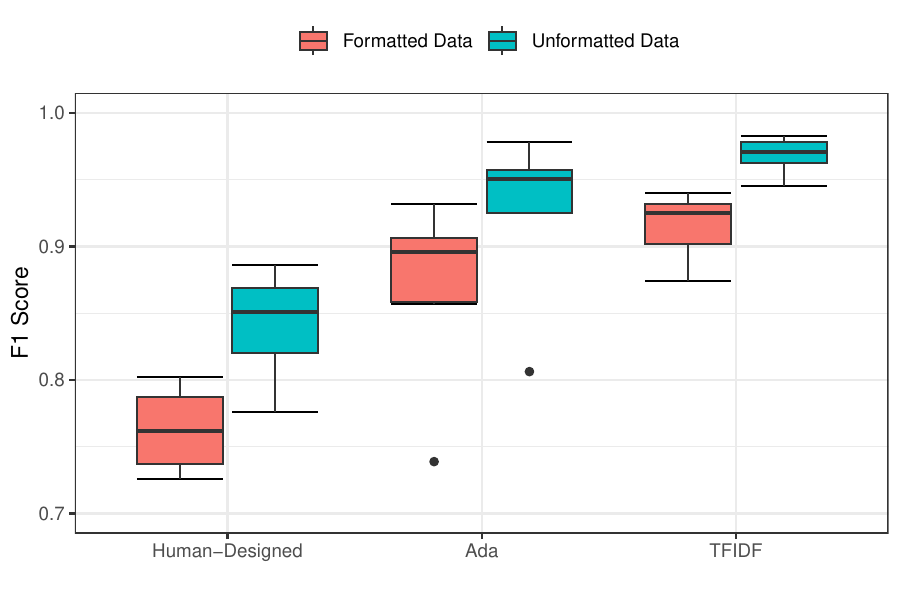}
    \caption{Distribution of the F1 Score of all considered algorithms in different conditions.
    In both \textsc{Ada} cases, the single outlier is CART.
    }
    \label{fig:Main_Results_Boxplot}
\end{figure}

\begin{table*}[tbp]
\caption{Same as Table~\ref{tab:evaluation_new_unformatted} but for \textbf{formatted} data.
}
\label{tab:evaluation_new_formatted}
\centering
\small
\begin{tabular}{llllll}
\toprule
& \multicolumn{5}{c}{Formatted} \\ \cmidrule(lr){2-6}
Model         & Accuracy ($\%$)    &   Precision ($\%$)   &   Recall ($\%$) &   F1 ($\%$)   &   AUC ($\%$)  \\
\midrule
& \multicolumn{5}{c}{Human-designed (white-box)} \\ \cmidrule(lr){1-6}
\textsc{CART}   &   $72.74 \pm 0.58$    &   $72.94 \pm 0.54$    &   $72.31 \pm 1.24$    &   $72.62 \pm 0.72$    &   $72.75 \pm 0.57$	\\
\textsc{DNN}    &   $77.51 \pm 0.89$    &   $78.15 \pm 3.65$    &   $77.18 \pm 6.72$    &   $77.31 \pm 2.33$    &   $86.69 \pm 0.30$    \\
\textsc{GMM}    &   $74.09 \pm 0.65$    &   $73.37 \pm 1.78$    &   $75.83 \pm 2.45$    &   $74.53 \pm 0.53$    &   $82.17 \pm 0.50$    \\
\textsc{LR}     &   $73.03 \pm 0.56$    &   $73.22 \pm 0.77$    &   $72.65 \pm 0.68$    &   $72.93 \pm 0.52$    &   $80.61 \pm 0.36$	\\
\textsc{OPCT}   &   $73.63 \pm 1.50$    &   $69.81 \pm 3.70$    &   $\mathbf{84.38 \pm 4.88}$    &   $76.18 \pm 0.85$    &   $80.78 \pm 1.76$	\\
\textsc{RF}     &   $\mathbf{80.50 \pm 0.43}$    &   $\mathbf{81.83 \pm 0.49}$    &   $78.42 \pm 0.69$    &   $80.09 \pm 0.47$    &   $\mathbf{88.86 \pm 0.30}$    \\	
\textsc{XGB}    &   $80.29 \pm 0.54$    &   $80.57 \pm 0.62$    &   $79.84 \pm 0.69$    &   $\mathbf{80.20 \pm 0.55}$    &   $88.65 \pm 0.35$	\\
\midrule
& \multicolumn{5}{c}{Embedding (black-box)} \\ 
\midrule
& \multicolumn{5}{c}{\textsc{Ada}} \\ \cmidrule(lr){1-6}
\textsc{CART}   &   $74.33 \pm 0.58$    &   $73.89 \pm 0.58$    &   $75.19 \pm 0.67$    &   $72.63 \pm 0.68$    &   $74.33 \pm 0.58$    \\
\textsc{DNN}    &   $\mathbf{93.14 \pm 0.87}$    &   $\mathbf{92.46 \pm 2.95}$    &   $\mathbf{94.12 \pm 2.21}$    &   $\mathbf{93.22 \pm 0.69}$    &   $\mathbf{98.66 \pm 0.10}$    \\
\textsc{GMM}    &   $86.10 \pm 0.40$    &   $88.16 \pm 0.55$    &   $83.41 \pm 0.55$    &   $85.71 \pm 0.41$    &   $93.03 \pm 0.31$	\\
\textsc{LR}     &   $91.16 \pm 0.35$    &   $91.21 \pm 0.35$    &   $90.72 \pm 0.48$    &   $91.70 \pm 0.56$    &   $97.29 \pm 0.17$    \\
\textsc{OPCT}   &   $89.99 \pm 0.40$    &   $90.03 \pm 0.41$    &   $89.67 \pm 1.42$    &   $90.44 \pm 1.72$    &   $94.56 \pm 0.86$    \\
\textsc{RF}     &   $85.89 \pm 0.34$    &   $85.94 \pm 0.37$    &   $85.63 \pm 0.59$    &   $86.26 \pm 0.89$    &   $94.25 \pm 0.18$    \\
\textsc{XGB}    &   $89.62 \pm 0.15$    &   $89.63 \pm 0.18$    &   $89.58 \pm 0.25$    &   $89.67 \pm 0.48$    &   $96.72 \pm 0.14$    \\  
\midrule
& \multicolumn{5}{c}{\textsc{TF-IDF}} \\ \cmidrule(lr){1-6}
\textsc{CART}   &   $89.69 \pm 0.41$    &   $89.65 \pm 0.43$    &   $90.01 \pm 0.43$    &   $89.30 \pm 0.67$    &   $89.69 \pm 0.41$    \\ 
\textsc{DNN}    &   $93.17 \pm 0.29$    &   $93.48 \pm 0.90$    &   $92.83 \pm 1.06$    &   $93.14 \pm 0.30$    &   $98.36 \pm 0.16$	\\
\textsc{GMM}    &   $87.53 \pm 0.52$    &   $88.14 \pm 0.61$    &   $86.74 \pm 0.67$    &   $87.43 \pm 0.53$    &   $91.01 \pm 0.58$    \\
\textsc{LR}     &   $92.50 \pm 0.36$    &   $92.52 \pm 0.37$    &   $92.32 \pm 0.37$    &   $92.72 \pm 0.58$    &   $97.95 \pm 0.22$    \\
\textsc{OPCT}   &   $90.74 \pm 0.57$    &   $90.78 \pm 0.53$    &   $90.48 \pm 1.61$    &   $91.12 \pm 1.51$    &   $94.76 \pm 0.64$    \\
\textsc{RF}     &   $93.36 \pm 0.37$    &   $93.29 \pm 0.38$    &   $\mathbf{94.33 \pm 0.36}$    &   $92.27 \pm 0.44$    &   $98.55 \pm 0.15$    \\
\textsc{XGB}    &   $\mathbf{93.98 \pm 0.44}$    &   $\mathbf{93.98 \pm 0.44}$    &   $93.96 \pm 0.51$    &   $\mathbf{94.01 \pm 0.48}$    &   $\mathbf{98.80 \pm 0.16}$    \\   
\midrule
& \multicolumn{5}{c}{\textsc{Word2Vec}} \\ \cmidrule(lr){1-6}
\textsc{GMM}    &   $87.92 \pm 0.45$    &   $87.61 \pm 0.82$    &   $88.35 \pm 0.95$    &   $87.97 \pm 0.45$    &   $89.26 \pm 0.41$    \\
\bottomrule
\end{tabular}
\end{table*}

\subsection{Gaussian Mixture Models}
\label{sec:gmm_results}
%
Although GMMs are, in principle, capable of unsupervised learning, they reach better classification accuracies in a supervised or semi-supervised setting: the class labels are provided during training, but the assignment of data points to one of the $K$ GMM components has to be found by the EM method described in Section~\ref{sec:gmm_methods}. Our method proceeds as follows:  
First, we train two independent GMMs, one exclusively on human samples and the other on GPT samples. For the prediction of an embedded sample $\vec{x}$, we then calculate the likelihood:
\begin{align*}
    p(\vec{x}; \mathcal{G}_\textsc{AI}) &= \sum_{k=1}^{K} \psi_k \mathcal{N} \left (\vec{x} | \vec{\mu}_k^\textsc{(AI)}, \Sigma_k^\textsc{(AI)} \right )  \\
    p(\vec{x}; \mathcal{G}_\textsc{HU}) &= \sum_{k=1}^{K} \psi_k \mathcal{N} \left (\vec{x} | \vec{\mu}_k^\textsc{(HU)}, \Sigma_k^\textsc{(HU)} \right )      
\end{align*}
where $p(\vec{x}; \mathcal{G}_\textsc{AI})$ and $p(\vec{x}; \mathcal{G}_\textsc{HU})$ denote the probability density functions of $\vec{x}$ under the GMMs, respectively, and assign $\vec{x}$ to the class with the higher probability.

When using \textsc{Ada}, the embedding can be performed on individual code snippets $x$ directly before the GMM finds clusters in the embeddings $\mathcal{E}(x)$. When applying TF-IDF embedding, instead, tokenization $\mathcal{T}$ is required (to determine the number of tokens in each snippet) before the embeddings are computed on the tokenized code snippets $\mathcal{E}(\mathcal{T}(x))$. 
With both embeddings, GMMs achieve accuracies of over $90\%$, outperforming any of the models based on human-designed features (see Table~\ref{tab:evaluation_new_unformatted}).

The underlying concept of GMMs is the approximation of the probability distributions of ChatGPT, which does not extend over entire snippets but rather on individual tokens used for the prediction of the following token. It appears, therefore, naturally, to apply Word2Vec for the embedding of single tokens instead of the snippet-level embedding used in \textsc{Ada} and TF-IDF.\footnote{It should be noted that while this approach would technically also work for \textsc{Ada}, the computational overhead makes it infeasible (as a rough estimate we consider $\approx 2.6 \times 10^{6}$ tokens and an average API response time of $\approx 5$ seconds).}
With the single-token Word2Vec embedding, GMMs reach an accuracy of $93.57\%$. 

To summarize, we show in Figure~\ref{fig:Main_Results_Boxplot} the box plots for all our main results. Each box group contains all models trained on a particular (feature set, format)-combination. Within a certain format choice, the box plots for the embedding feature sets do not overlap with those for the human-designed feature set. This shows that the choice `embedding vs. human-designed' is more important than the specific ML model.


\begin{figure}[h!]
    \centering
    \hspace*{-4cm}
    \includegraphics[width=1.3\linewidth]{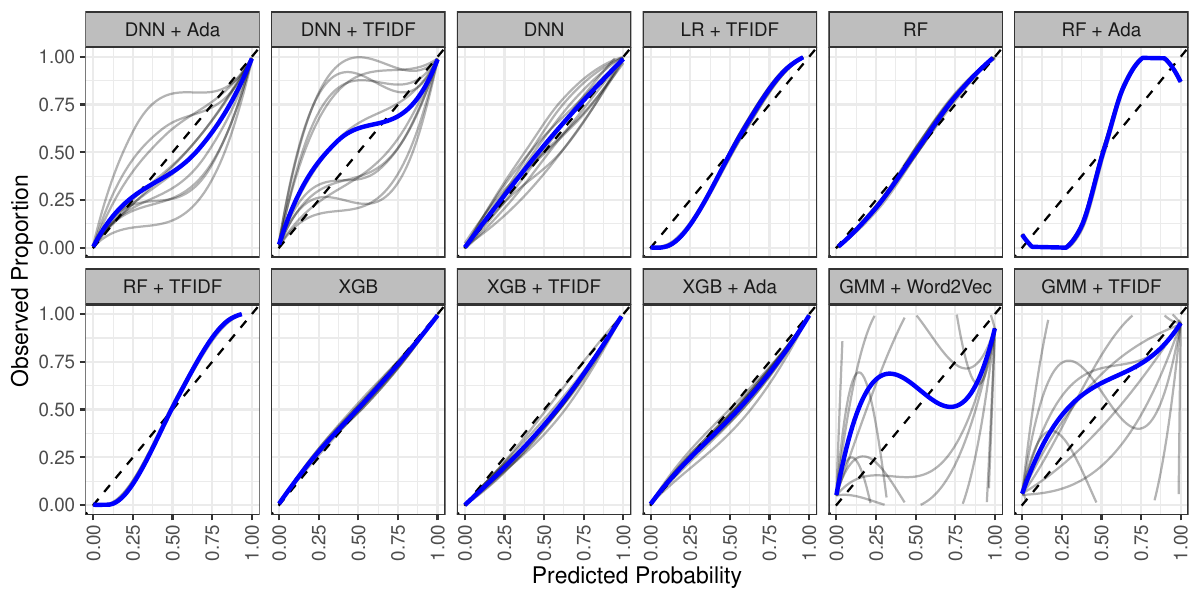}
    \caption{Calibration plots for selected 
    classifiers trained and evaluated using \textbf{unformatted} test-set data. The lines were estimated using LOESS regression models (gray: individual runs, blue: mean over all runs). 
    Dashed diagonal line: perfect calibration.}
    \label{fig:unform_calibration}
\end{figure}

\begin{figure}[h!]
    \centering
    \hspace*{-4cm}
    \includegraphics[width=1.3\linewidth]{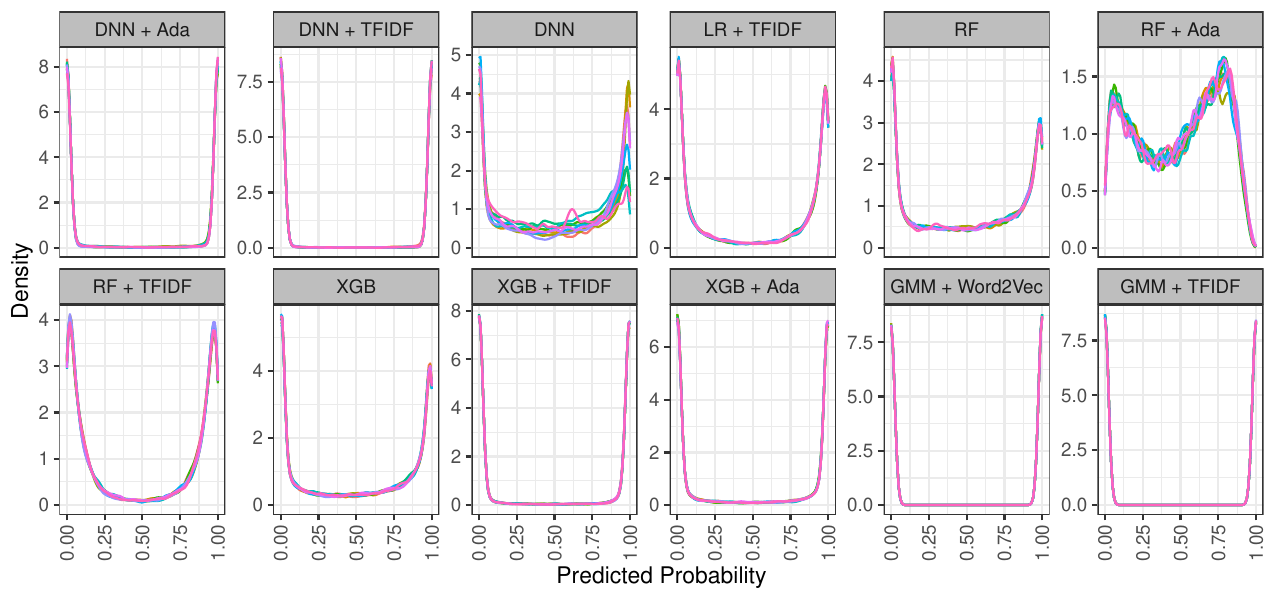}
    \caption{Kernel density estimates of the probabilities predicted by each considered classifier in the \textbf{unformatted} test set. The plot contains one line per run. }
    \label{fig:unform_pred_dens}
\end{figure}

\begin{figure}[h!]
    \centering
    \hspace*{-4cm}
    \includegraphics[width=1.3\linewidth]{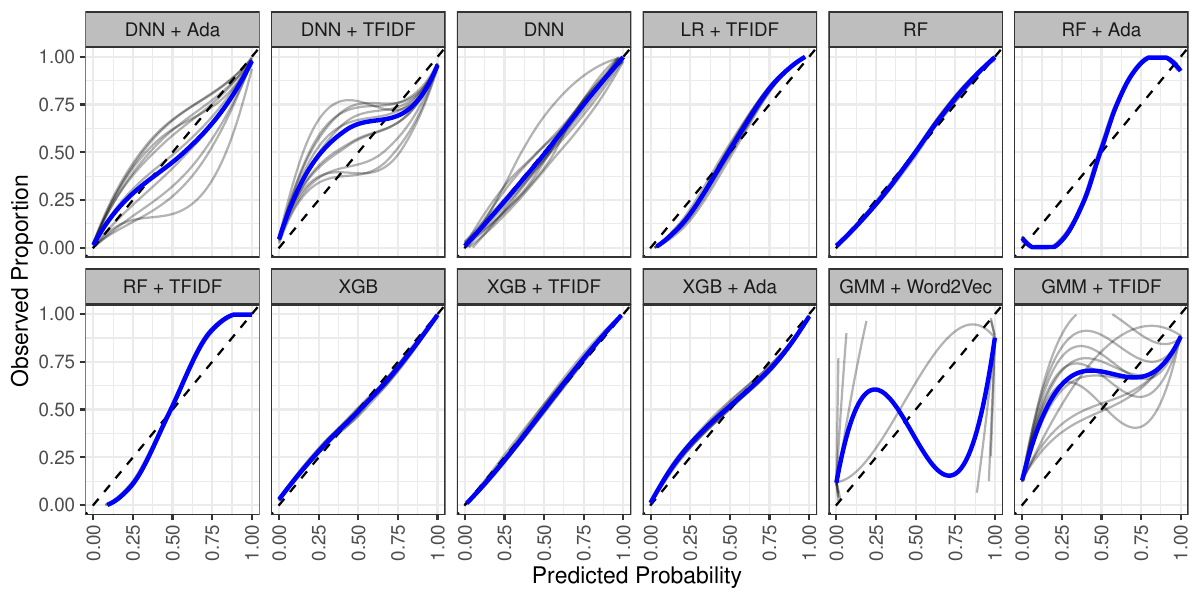}
    \caption{Calibration plots for selected 
    classifiers trained and evaluated using \textbf{formatted} test-set data. The lines were estimated using LOESS regression models (gray: individual runs, blue: mean over all runs). 
    Dashed diagonal line: perfect calibration.}
    \label{fig:form_calibration}
\end{figure}

\begin{figure}[h!]
    \hspace*{-4cm}
    \includegraphics[width=1.3\linewidth]{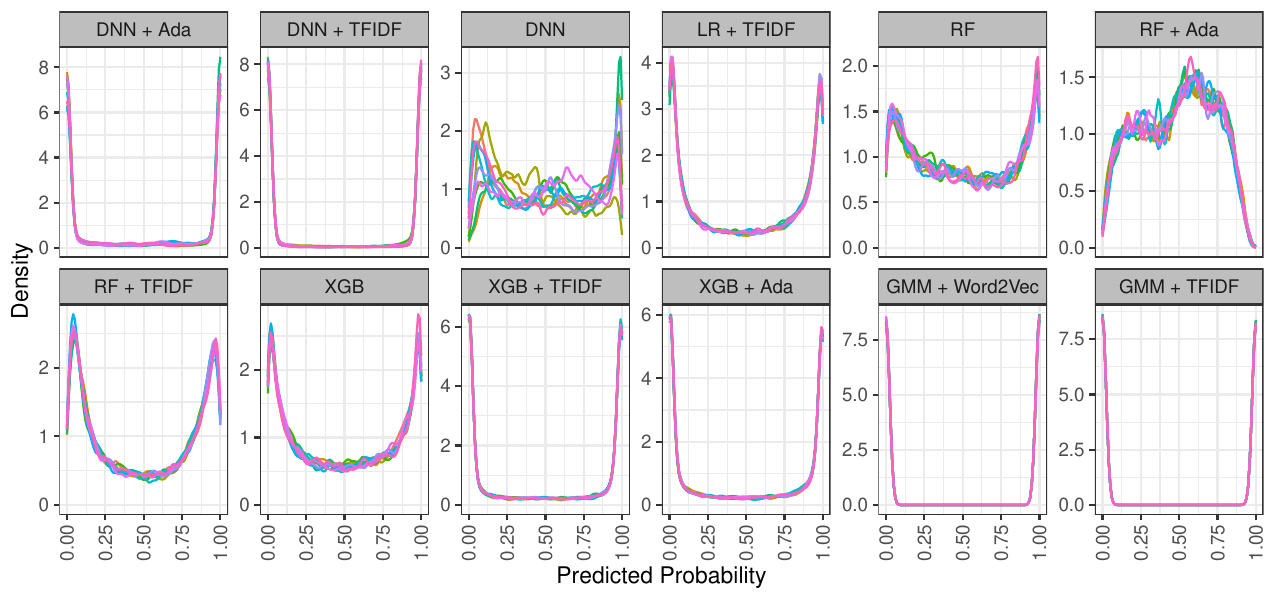}
    \caption{Kernel density estimates of the probabilities predicted by each considered classifier in the \textbf{formatted} test set. The plot contains one line per run.}
    \label{fig:form_pred_dens}
\end{figure}

\subsection{Model Calibration}
\label{sec:model_calib}
Previously, we focused on how well the considered algorithms can discriminate between code snippets generated by humans and GPT-generated code snippets. This is an important aspect when judging the performance of these algorithms. However, it also reduces the problem to one of classification, e.g., the algorithm only tells us whether a code snippet is generated by humans or by GPT. In many cases, we may also be interested in more nuanced judgments, e.g., in \emph{how likely} it is that GPT or humans generated a code snippet. All the considered algorithms are theoretically able to output such predicted class probabilities. In this section, we evaluate how well these predicted probabilities correspond to the proportion of actually observed cases.

We mainly use calibration plots to do this for a subset of the considered algorithms. In such plots, the predicted probability of a code snippet being generated by GPT is shown on the $x$-axis, while the actually observed value ($0$ if human-generated, $1$ if GPT generated) is shown on the $y$-axis. A simple version of this plot would divide the predicted probabilities into categories, for example, ten equally wide ones~\cite{broecker2007}. The proportion of code snippets labeled as GPT inside those categories should ideally be equal to the mean predicted probability inside this category. For example, in the category $10-20\%$, the proportion of actual GPT samples should roughly equal $15\%$. We use a smooth variant of this plot by calculating and plotting a non-parametric locally weighted regression (LOESS) instead, which does not require the use of arbitrary categories~\cite{austin2014}.

Figures~\ref{fig:unform_calibration} and~\ref{fig:form_calibration} show these calibration plots for each algorithm, separately for formatted and unformatted data. Most of the considered algorithms show adequate calibration, with the estimated LOESS regression being close to the line that goes through the origin. There are only minor differences between algorithms fitted on formatted vs unformatted data. For some algorithms, however (\textsc{RF + Ada}, \textsc{RF + TF-IDF}, \textsc{GMM + Word2Vec}, \textsc{GMM + TF-IDF}), there seem to be some issues with the calibration in predicted probabilities between $0.15$ and $0.85$. However, one possible reason for this is not a lack of calibration but a lack of suitable data points to correctly fit the LOESS regression line. For example, the GMM-based models almost always predict only probabilities very close to $0$ or $1$. This is not necessarily a problem of calibration (if the algorithm is correct most of the time), but it may lead to unstable LOESS results~\cite{austin2014}.

We, therefore, additionally plotted kernel density estimates of the predicted probabilities for each algorithm to show the range of predictions made by each one (Figures~\ref{fig:unform_pred_dens} and~\ref{fig:form_pred_dens}). As can be seen quite clearly, most of the DNN and GMM-based models relying on embeddings generated only very few predicted probabilities between $0.1$ and $0.9$, making the validity of the calibration curves in these ranges questionable for those algorithms. Since these models do not discriminate perfectly between the two classes, these models may not be the best choice when the main interest lies in predicting the probability of a code snippet being generated by GPT because their output always suggests certainty, even when it is wrong. On the other hand, algorithms such as \textsc{XGB + TF-IDF} or \textsc{XGB + Ada} show a nearly perfect calibration and a very high accuracy.

\begin{figure}[h]
    \centering
    \includegraphics[width=0.7\textwidth]{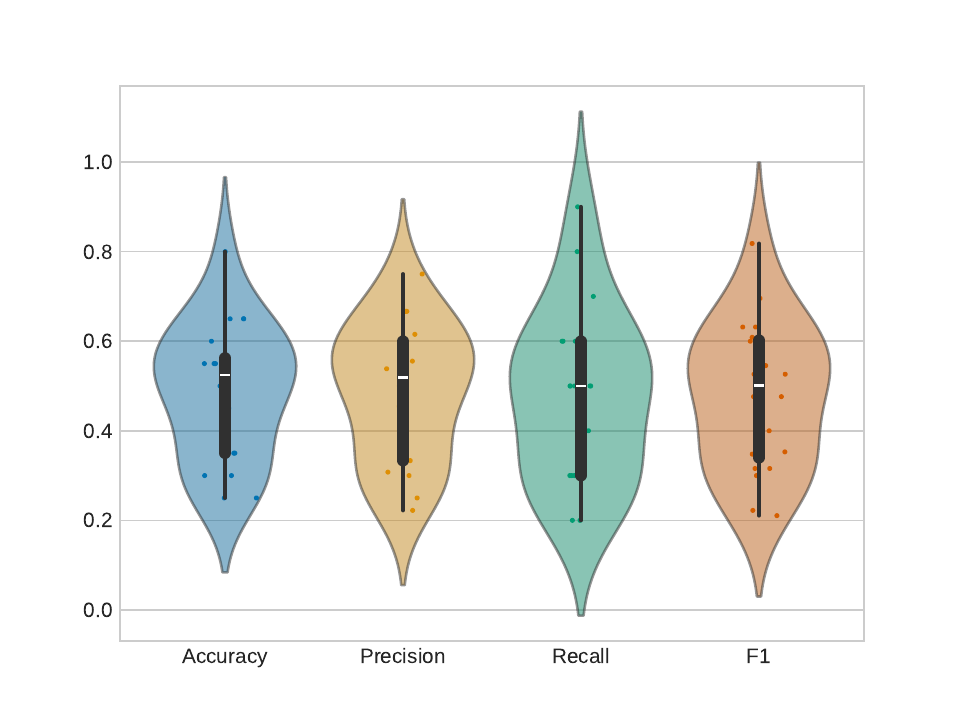}
    \caption{Performance distribution of study participants 
    }
    \label{fig:survey}
\end{figure}

\subsection{Human Agents and Bayes Classifiers}
To gain a better understanding about the ``baseline'' performance, to which we compare the more sophisticated models, we briefly present the results of untrained humans and a Bayes classifier.

\subsubsection{Untrained Human Agents}
\label{subsubsec:human_agents}
To assess how difficult it is for humans to classify Python code snippets according to their origin, we conducted a small study with $20$ participants. Using Google Forms, the individuals were asked to indicate their educational background, their experience in Python, and to self-asses their programming proficiency on a $10$-point Likert scale. They were afterward asked to judge whether $20$ randomly selected and ordered code snippets were written by humans or by ChatGPT. The dataset was balanced, but the participants were not told.
The participants, of which $50\%$ had a Master's degree and $40\%$ held a PhD degree, self-rated their programming skills at an average of $6.85 \pm 1.69$ with a median of $7$ and indicated an average of $ 5.05 \pm 4.08$ (median $5$) years of Python programming experience.
The result in Table~\ref{tab:survey} shows a performance slightly below random guessing, confirming this task's difficulty for untrained humans. Figure~\ref{fig:survey} shows the distribution of the participant's performances.
\begin{table*}[h!]
    \vspace{0.1cm}
    \caption{Results of human performance on the classification task}
    \label{tab:survey}
    \centering
        \begin{tabular}{llll}
        \toprule
        Accuracy (\%) & Precision (\%) & Recall (\%) & F1 (\%) \\
        \midrule
        $48.7 \pm 14.7$ & $48.4 \pm 14.8$ & $48.0 \pm 18.9$ & $47.6 \pm 16.0$ \\
        \bottomrule
        \end{tabular}
\end{table*}

When comparing these results to the ones obtained by ML models, it should be taken into account that the participants were not trained on the task and are, therefore, considered zero-shot learners. How well humans can be trained in the task using labeled data remains to be investigated.

\begin{figure}
    \centering
    \includegraphics[width=\textwidth]{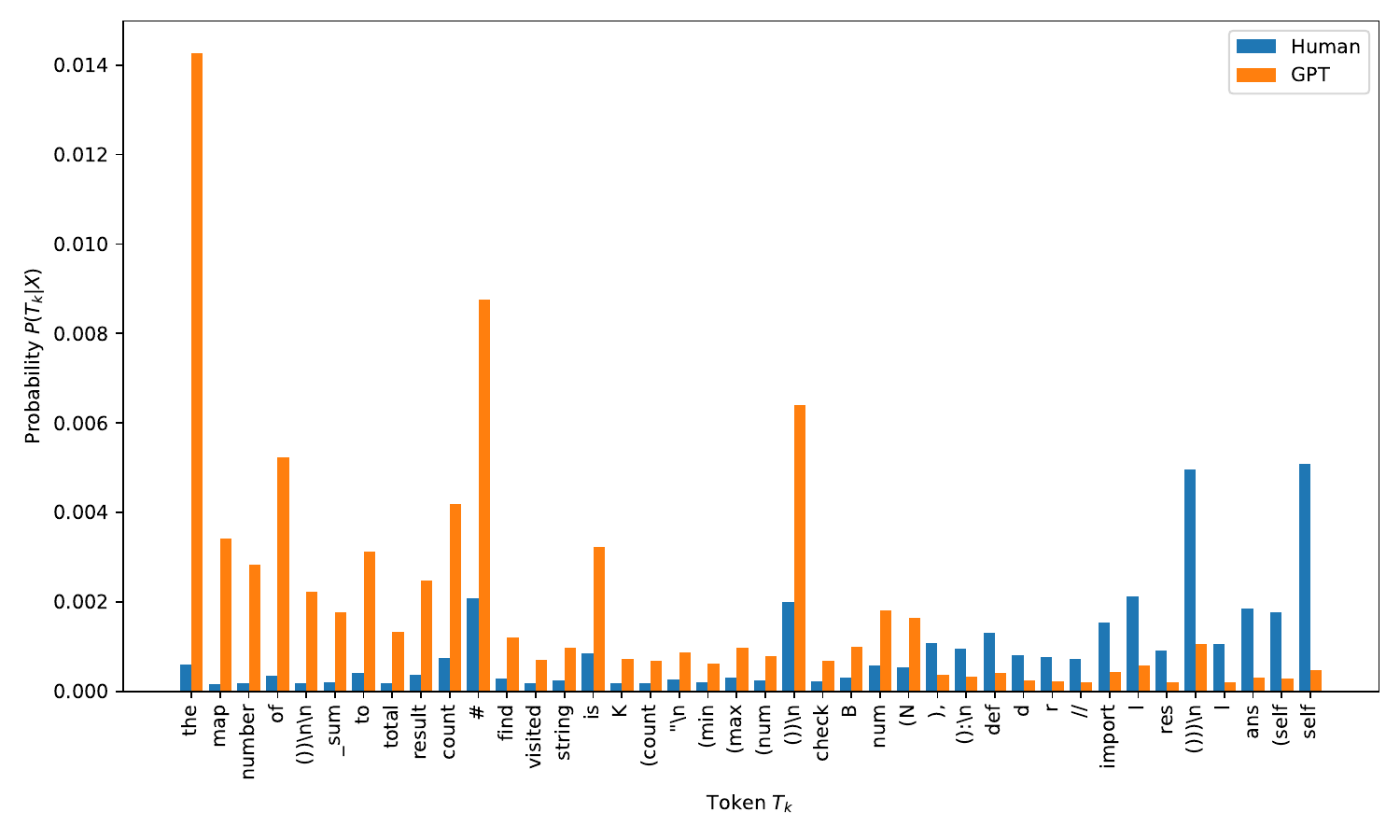}
    \caption{Top $40$ tokens with the largest absolute discrepancies in their log probabilities (which corresponds to the largest ratio of probabilities).}
    \label{fig:Filtered_Top_40_Token_Log_Discrepancies}
\end{figure}


\subsubsection{Trained Bayes Classifier}
\label{sec:bayes_classif}
%
%
Viewed from a broader perspective, the question raised by the results from the preceding subsection is why ML models with an F1 score between $83\%$ and $98\%$ excel so much over (untrained) humans who show a performance close to random guessing. 
In the following, we investigate the possibility (which also appeared briefly in \citeauthor{li2023discriminating}~\cite{li2023discriminating}) that ML models might use subtle differences in conditional probabilities for the appearance of tokens for decision-making. Such probabilities might be complex for humans to calculate, memorize, and combine.

To demonstrate this, we build a simple Bayes classifier: In the following, we abbreviate $H =$ human origin, $G =$ GPT origin, and $X =$ either origin. For each token $t_k$ in our training dataset, we calculate the probabilities $P(t_k|X)$. We keep for reliable estimates only those tokens $t_k$ where the absolute frequencies $n(t_k|H)$ and $n(t_k|G)$ are both greater-equal some predefined threshold $\tau$. The set of those tokens above the threshold shall be $T$.
For illustration, Figure~\ref{fig:Filtered_Top_40_Token_Log_Discrepancies} shows the tokens with the largest probability ratio.

Given a new code document $D$ from the test dataset, we determine the intersection $D \cap T$ of tokens and enumerate the elements in this intersection as $\{T_1, \ldots, T_K\}$, where $T_k$ means the event ``Document $D$ contains token $t_k$''. We assume statistical independence:
$P(\cap_{k=1}^K T_k|X) = \prod_{k=1}^K P(T_k |X).$
Now we can calculate with Bayes's law the probabilities of origin:
\begin{align}
P(H|\cap_{k=1}^K T_k) &= \frac{(P(\cap_{k=1}^K T_k |H)P(H))}
    {(P(\cap_{k=1}^K T_k |H)P(H)+P(\cap_{k=1}^K T_k |G)P(G))} \notag \\                     &= \frac{\left[\prod_{k=1}^K P(T_k  |H)\right]P(H))}
                    {\left[\prod_{k=1}^K P(T_k|H)\right]P(H) + 
                     \left[\prod_{k=1}^K P(T_k |G)\right]P(G)}  \label{eq:Bayes}
\end{align}
and, similarly, $P(G|\cap_{k=1}^K T_k)$. By definition, we have $P(H|\cap_{k=1}^K T_k)+P(G|\cap_{k=1}^K T_k)=1$.

The Bayes classifier classifies document $D$ as being of GPT-origin, if $P(G|\cap_{k=1}^K T_k) > P(H|\cap_{k=1}^K T_k)$.
Eq.~\eqref{eq:Bayes} might look complex, but it is just a simple multiply-add of probabilities, straightforward to calculate from the training dataset. It is a simple calculation for a machine but difficult for a human just looking at a code snippet. 

We conducted the Bayes classifier experiment with $\tau=32$ (best out of a number of tested $\tau$ values), and obtained the results shown in Table~\ref{tab:bayes}.
\begin{table*}[tbp]
    \vspace{0.1cm}
    \caption{Results of Bayes classifier. Mean and standard deviation from $10$ runs with $10$ different training-test-set separations (problem-wise).}
    \label{tab:bayes}
    \centering
        \begin{tabular}{l|llll}
        \toprule
        &Accuracy (\%) & Precision (\%) & Recall (\%) & F1 (\%) \\
        \midrule
        Unformatted & $87.77 \pm 0.56$  &   $82.99 \pm 0.70$    &   $95.03 \pm 0.58$    &   $88.60 \pm 0.50$ \\
        Formatted   & $86.06 \pm 0.65$  &   $85.85 \pm 1.23$    &   $86.46 \pm 1.02$    &   $86.15 \pm 0.59$ \\
        \bottomrule
        \end{tabular}
\end{table*}

It is astounding that such a simple ML model can achieve results comparable to the best human-designed-features (white-box) models (unformatted, Table~\ref{tab:evaluation_new_unformatted}) or the average of the GMM models (formatted, Table~\ref{tab:evaluation_new_formatted}). On the other hand, the Bayes calculation is built upon a large number of features (about $1366$ tokens in the training token set $T$ and on average $144$ tokens per tested document), which makes it clear that a (untrained) human cannot perform such a calculation by just looking at the code snippets. This is notwithstanding the possibility that a trained human could somehow learn an equivalent complex pattern matching that associates the presence or absence of specific tokens with the probability of origin.

\section{Discussion}
\label{sec:Discussion}
In this section, we will explore the strengths and weaknesses of the methodologies employed in this paper, assess the practical implications of our findings, and consider how emerging technologies and approaches could further advance this field of study. Additionally, we compare our results with those of other researchers attempting to detect the origin of code.

\subsection{Strengths and Weaknesses}
\label{sec:strength_weak}

\begin{itemize}
    \item[] \textbf{Large code dataset}: To the best of our knowledge, this study currently is the largest in terms of coding problems collected and solved by both humans and AI (see Table~\ref{tab:total_samples}, $3.14\times 10^4$ samples in total). 
    The richness of the training data is probably responsible for the high accuracy and recall around $98\%$ we achieve with some of the SL models.
    \vspace{0.1cm} 
    \item[] \textbf{Train-test-split along problem instances}: A probably important finding from our research is that we initially reached a somewhat higher accuracy ($+4$ percentage points for most models, not shown in the tables above) with random-sample split. But this split method is flawed, if problems have many human solutions or many GPT solutions, because the test set may contain problems already seen in the training set. A split along problem instances, as we do in our final experiments, 
    ensures that each test case comes from a so-far unseen problem and
    is the choice recommended to other researchers as well. It has a lower accuracy, but it is the more realistic accuracy we expect to see on genuinely new problem instances. 
    \vspace{0.1cm} 
    \item[] \textbf{Formatting}: 
    Somewhat surprisingly, after subjecting all samples to the Black code formatter tool~\cite{black}, 
    our classification models still exhibited the capacity to yield satisfactory outcomes, only slightly degrading in performance. This shows a remarkable robustness. However, it is essential to note that the Black formatter represents just one among several code formatting tools available, each with its distinct style and rules. Utilizing a different formatter such as AutoPEP8~\cite{autopep} could potentially introduce variability in the formatting of code snippets, impacting the differentiation capability of the models. Examining the robustness of our classification models against diverse formatting styles remains an area warranting further exploration. Additionally, the adaptation of models to various formatters can lead to the enhancement of their generalization ability, ensuring consistent performance across different coding styles.

   
   \vspace{0.1cm} 
    \item[] \textbf{White-box vs black-box features}: Our experiments on unformatted code have shown that the human-designed white-box features of Table~\ref{tab:exp_features} can achieve a good accuracy level above $80\%$ with most ML models. However, our exceptional good results of $92-98\%$ are only achieved with black-box embedding features. This is relatively independent of the ML model selected; what is more important is the type of input features.
    
   \vspace{0.1cm} 
    \item[] \textbf{Feature selection}: 
    Although the human-designed features (white-box) were quite successful for both the original and formatted code snippets, there is still a need for features that lead to higher accuracy. The exemplary performance of the embedding features (black-box) indicates the existence of features with higher discriminatory power, thereby necessitating a more profound analysis of the embedding space. 
    %
    
    
    \vspace{0.1cm} 
    \item[] \textbf{Bayes Classifier}: The Bayes classifier introduced in Section~\ref{sec:bayes_classif} shows another possibility to generate a rich and interpretable feature set. The statistical properties that can be derived from the training set enable an explainable classifier that achieves an accuracy of almost $90\%$. 
    
    \vspace{0.1cm} 
    \item[] \textbf{Test cases}: Tested code, having undergone rigorous validation, offers a reliable and stable dataset, enhancing the model’s accuracy and generalization by mitigating the risk of incorporating errors or anomalies. 
    This reliability fosters a robust training environment, enabling the model to learn discernible patterns and characteristics intrinsic to human- and AI-generated code. However, focusing solely on tested code may limit the model's exposure to diverse and unconventional coding styles or structures, potentially narrowing its capability to distinguish untested, novel, or outlier instances. Integrating untested code could enrich the diversity and comprehensiveness of the training dataset, accommodating a broader spectrum of coding styles, nuances, and potential errors, thereby enhancing the model's versatility and resilience in varied scenarios. In future work, exploring the trade-off between the reliability of tested code and the diversity of untested code might be beneficial to optimize the balance between model accuracy and adaptability across various coding scenarios.
    
    \vspace{0.1cm} 
    \item[] \textbf{Code generators}: 
    It is important to note that only AI examples generated with Open\-AI's \texttt{gpt-3.5-turbo} API were used in this experiment.
    This model is probably the one most frequently used by students, 
    rendering it highly suitable for the objectives of this research paper. However, it is not the most capable of the GPT series. Consequently, the integration of additional models such as \texttt{gpt-4}~\cite{openai2023gpt4} or \texttt{T5}$^{+}$~\cite{wang2023codet5}, which have demonstrated superior performance in coding-related tasks, can serve to not only enhance the utilization of the available data but also introduce increased variability. 
    It is an open research question whether \textit{one} classification model can disentangle several code generators and human code or whether separate models for each code generator are needed. 
    Investigating whether distinct models formulate their unique distributions or align with a generalized AI distribution would be intriguing. This exploration could provide pivotal insights into the heterogeneity or homogeneity of AI-generated code, thereby contributing to the refinement of methodologies employed in differentiating between human- and AI-generated instances.
    
    \vspace{0.1cm} 
    \item[] \textbf{Programming languages}: Moreover, it is essential to underscore that this experiment exclusively encompassed Python code. However, our approach remains programming-language-agnostic: Given a code dataset in another programming language, the same methods shown here for Python could be used to extract features or to embed the code snippets (token-wise or as a whole) in an embedding space. 
    We suspect that, given a similar dataset, the performance of classifiers built in such way would be comparable to the ones presented in this article.
    %

    \vspace{0.1cm}
    \item[] \textbf{Publicly available dataset and trained models}: To support further research on more powerful models or explainability in detecting AI-generated code, we made the pre-processed dataset publicly available in our repository \url{https://github.com/MarcOedingen/ChatGPT-Code-Detection}. The dataset can be downloaded via a link from there. Moreover, we offer several trained models and a demo version of the XGB model using the TF-IDF embedding. This demonstration serves as a counterpart to public AI-text detectors, allowing for the rapid online classification of code snippets. 
\end{itemize}


\subsection{Comparison with Other Approaches to Detect the Source of Code}

Two of the works mentioned in Section~\ref{sec:related}, namely \citeauthor{Hoq_Code_Detection}~\cite{Hoq_Code_Detection} and \citeauthor{yang2023zeroshot}~\cite{yang2023zeroshot}, are striving for the same goal as our paper: the detection of the source of code. Here, we compare their results with ours.

\citeauthor{Hoq_Code_Detection}~\cite{Hoq_Code_Detection} approach the source-of-code detection with two ML models (SVM, XGB) and with two DL models (code2vec, ASTNN). They find that all models deliver quite similar accuracies in the range of $90-95\%$. The drawback is, however, that they have a dataset with only $10$ coding problems, for each of which they generate $300$ solutions. They describe it as \textit{``limiting the variety of code structures and syntax that ChatGPT would produce''}. Moreover, given this small number of problems, a potential flaw is that a purely random train-test-split will have a high probability that each test problem is also represented in the training set (overfitting). In our approach with the larger code dataset, we perform the train-test-split in such a way that all test problems do not occur in the training set. This somewhat tougher task may be the reason for seeing a larger gap between simple ML models and more complex DL-embedding models than is reported in~\cite{Hoq_Code_Detection}. 

\citeauthor{yang2023zeroshot}~\cite{yang2023zeroshot} pursue the ambitious task of zero-shot classification, i.e., predicting the source of code without training. Even more ambitious, they use three different advanced LLMs as code generators (and not only ChatGPT3.5 as we do). Their number of code samples for testing is $267$, which is pretty low. Nothing is said about the number of AI-generated samples. Their specific method, sketched in Section~\ref{sec:related}, is shown in~\cite{yang2023zeroshot} to be much better than text detectors applied to the code detection task. However, without specific training, their TPR (= recall) of $20-60\%$ is much lower than the recall of $98\%$ we achieve with the best of our trained models.


\section{Conclusion}
\label{sec:Conclusion}
This research aimed to find a classification model capable of differentiating AI- and human-written code samples. 
In order to enable the feasibility of such a model in an application, emphasis was also put on explainability. Upon thoroughly examining existing AI-based text sample detection research, we strategically transposed the acquired knowledge to address the novel challenge of identifying AI-generated code samples.

We experimented with a variety of feature sets and a large number of ML models, including DNNs and GMMs. It turned out that the choice of the input feature set was more important than the model used. The best combinations are \textsc{DNN + Ada} with $97.8\%$ accuracy and \textsc{XGB + TF-IDF} with $98.3\%$ accuracy. The accuracies with human-designed, low-dimensional feature sets are $10-15$ percentage points lower.

Within the structured context of our experimental framework, and through the application of the methodologies and evaluative techniques delineated in this manuscript, we have successfully demonstrated the validity of our posited hypotheses H$_\mathbf{1}$ and H$_\mathbf{2}$ 
which postulate the distinctness between AI-generated and human coding styles, regardless of formatting. 
The interpretability of our approach was improved by a Bayes classifier, which made it possible to highlight individual tokens and provide a more differentiated understanding of the decision-making process.

One notable outcome of this experiment is the acquisition of a substantial dataset of Python code examples created by humans and AI, originating from various online coding task sources. This dataset serves as a valuable resource for conducting in-depth investigations into the fundamental structural characteristics of AI-generated code, and it facilitates a comparative analysis between AI-generated and human-generated code, highlighting distinctions and similarities. 

Having only experimented with Python code, further research could be focused on investigating the coding style of AI with other programming languages. Moreover, AI code samples were only created with OpenAI's \texttt{gpt-3.5-turbo} API. To make a more general statement about the coding style of language models, further research on other models should be conducted. This study, one of the first of its kind, can be used as a foundation for further research to understand how language models write code and how it differs from human-written code.

In light of the rapid evolution and remarkable capabilities of recent language models, which, while designed to benefit society, also harbor the potential for malicious use, developers and regulators must implement stringent guidelines and monitoring mechanisms to mitigate risks and ensure ethical usage. To effectively mitigate the potential misuse of these advances, the continued development of detection applications, informed by research such as that presented in this paper, remains indispensable.


\vspace{6pt} 

\authorcontributions{
Conceptualization, M.O. and M.H.; 
methodology, M.O., M.H. and W.K.; 
software, M.O., R.D. and M.H.; 
validation, M.O., R.C.E., R.D. and W.K.;
formal analysis, M.O., R.C.E., R.D. and W.K.;
investigation, M.O., R.C.E., R.D. and W.K.;
resources, M.O., R.C.E. and M.H.; 
data curation, M.O. and M.H.; 
writing---original draft preparation, M.O., R.C.E., R.D., M.H. and W.K.; 
writing---review and editing, M.O., R.C.E., R.D. and W.K.;
visualization, M.O., R.C.E., R.D. and W.K.; 
supervision, M.O. and W.K.; 
project administration, M.O. and W.K. 
All authors have read and agreed to the published version of the manuscript.
}

\abbreviations{Abbreviations}{
The following abbreviations are used in this manuscript: 
\\

\noindent 
\begin{tabular}{@{}ll}
AUC & Area Under the Curve (ROC curve) \\
ASTNN & Abstract syntax tree-based neural networks\\
CART & Classification and Regression Tree \\
CBOW & Continuous bag of words \\
DNN & Deep Neural Network\\
DT & Decision Tree \\
EM & Expectation maximization \\
F1 & $F_1$-score\\ 
GMM & Gaussian Mixture Model\\
GPT & Generative Pretrained Transformer\\
LLM & Large Language Model \\
LR & Logistic Regression \\
LSTM & Long-short-term memory \\
ML & Machine Learning\\
NL & Natural Language\\
OPCT & Oblique Predictive Clustering Tree \\
RF & Random Forest \\
SL & Supervised Learning\\
TF-IDF & Term frequency-inverse document frequency\\
XGB & eXtreme Gradient Boosting \\
\end{tabular}
}

\appendixtitles{no} 
\appendixstart
\appendix

\begin{adjustwidth}{-\extralength}{0cm}

\reftitle{References}


\bibliography{literature.bib}

\PublishersNote{}
\end{adjustwidth}


\end{document}